\documentclass[letterpaper,twocolumn,10pt]{article}
\usepackage{usenix-2020-09}

\usepackage{tikz}
\usepackage{amsmath}

\usepackage{filecontents}

\usepackage{amsmath}
\usepackage{amssymb}
\usepackage{bbm}
\usepackage{algpseudocode}
\usepackage[ruled,linesnumbered]{algorithm2e}
\usepackage{subfig}
\usepackage{booktabs}
\usepackage{multirow}
\usepackage{float}
\usepackage[section]{placeins}

\begin{document}

\date{}

\title{\Large \bf Forgetting Through Transforming: Enabling Federated Unlearning via Class-Aware Representation Transformation}


\author{
{\rm Qi Guo\footnotemark[1], Zhen Tian\footnotemark[1], Minghao Yao, Yong Qi, Saiyu Qi}\\
Xi'an Jiaotong University \\
qiguoqg@outlook.com
\and
{\rm Yun Li}\\
Shanghai Jiao Tong University   \\
lin\_yun@sjtu.edu.cn
\and
{\rm Jin Song Dong}\\
National University of Singapore  \\
dcsdjs@nus.edu.sg
} 

\maketitle
\def\thefootnote{$*$}\footnotetext{Equal contribution}
\begin{abstract}
Federated Unlearning (FU) enables clients to selectively remove the influence of specific data from a trained federated learning model, addressing privacy concerns and regulatory requirements. However, existing FU methods often struggle to balance effective erasure with model utility preservation, especially for class-level unlearning in non-IID settings. We propose Federated Unlearning via Class-aware Representation Transformation (FUCRT), a novel method that achieves unlearning through class-aware representation transformation. FUCRT employs two key components: (1) a transformation class selection strategy to identify optimal forgetting directions, and (2) a transformation alignment technique using dual class-aware contrastive learning to ensure consistent transformations across clients. Extensive experiments on four datasets demonstrate FUCRT's superior performance in terms of erasure guarantee, model utility preservation, and efficiency. FUCRT achieves complete (100\%) erasure of unlearning classes while maintaining or improving performance on remaining classes, outperforming state-of-the-art baselines across both IID and Non-IID settings. Analysis of the representation space reveals FUCRT's ability to effectively merge unlearning class representations with the transformation class from remaining classes, closely mimicking the model retrained from scratch.
\end{abstract}
\section{Introduction}
Federated Learning (FL) is an emerging and promising distributed learning paradigm, allowing multiple clients to jointly train a global model without directly sharing their raw private data \cite{mcmahan2017communication}. 
In FL, clients train on an initial model locally with their own data, and upload their model updates to the federated server to build a shared global model.
An essential requirement within FL is the clients’ “right to be forgotten”, as explicitly outlined in regulations such as the European Union General Data Protection Regulation (GDPR) \cite{voigt2017eu} and the California Consumer Privacy Act (CCPA) \cite{harding2019understanding}, which give clients the right to eliminate specific categories of data from the trained model. 
To address this requirement, Federated Unlearning (FU) has been introduced, enabling clients to selectively remove the influence of specific subsets of their data from a trained FL model \cite{wu2022federated,wang2022federated,zhao2023federated,yang2023survey}. Through the application of FU, organizations can uphold privacy regulations, protect user information, and maintain the performance of FL systems on the remaining data \cite{liu2023survey}.

\begin{figure}[t]
  \centering
  \subfloat[Original model]{
    \includegraphics[width=0.24\textwidth]{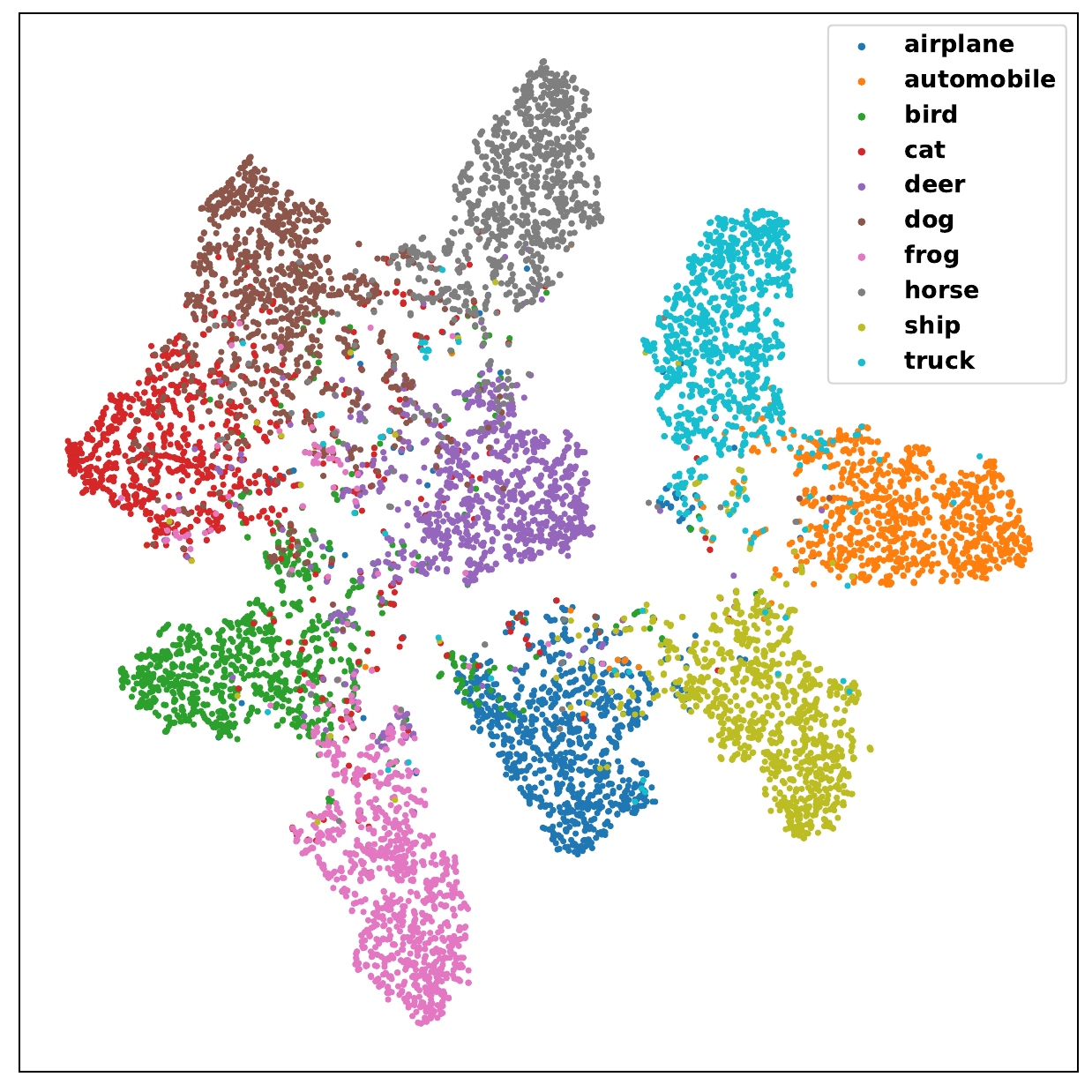}
  }
  \subfloat[Retained from Scratch model]{
    \includegraphics[width=0.24\textwidth]{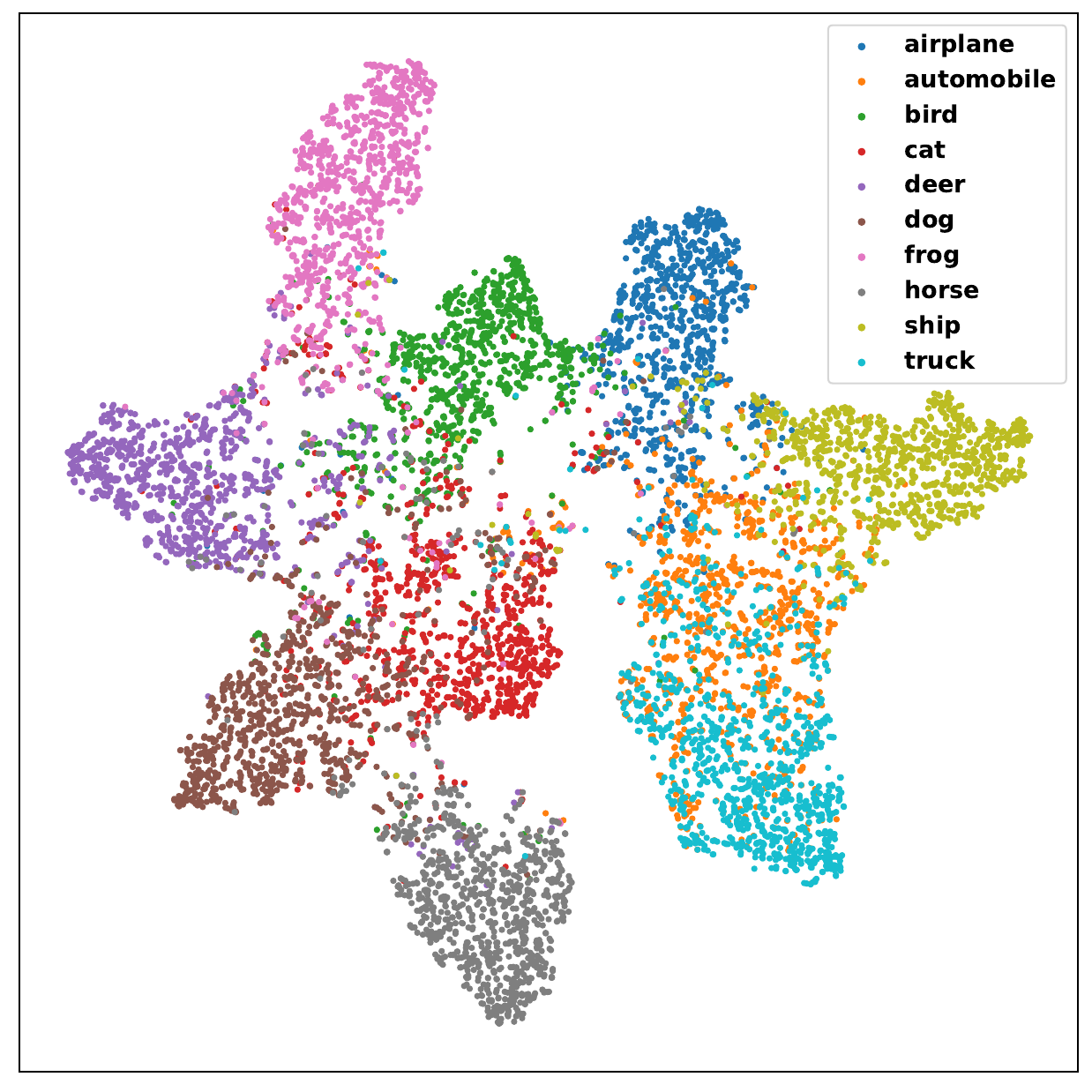}
  }
  \caption{Distribution of different classes in the model's representation space for the CIFAR10 dataset. The class “automobile” shown in orange, represents the unlearning class. It can be observed that: (1) The representations of the remaining class data exhibit compact and separable clustering characteristics in the representation space of both the original model and the model retrained from scratch. (2) The representations of the unlearning class data (i.e., “automobile”) are concentrated within specific representation domains associated with certain classes (i.e., “truck” and “airplane”) in the representation space of the model retrained from scratch.}
  \label{Fig:introduction}
\end{figure}

In this paper, we focus on unlearning data of specific target classes from a trained FL model, which is useful in realistic scenarios like face recognition, diagnosis records, and bank statements. A naive way to achieve FU is to retrain from scratch. However, the computational and time costs incurred by fully retraining models following data erasure can be prohibitively high, making this approach impractical in real-world scenarios. The existing works can be divided into three main categories: the reverse stochastic gradient ascent \cite{wu2022federated}, the channel pruning \cite{wang2022federated}, and momentum degradation \cite{zhao2023federated}. The gradient-ascent method uses the reverse stochastic gradient ascent to tighten the generalization boundary and effectively eliminate the model's capacity to classify specific data \cite{wu2022federated}. Leveraging term frequency-inverse document frequency, FUDP seeks to prune the most relevant channels associated with the target classes, effectively unlearning their contributions to the model \cite{wang2022federated}. FUMD utilizes an isomorphic and randomly initialized degradation model to erase the existence of the unlearning class on the target model \cite{zhao2023federated}. These works tend to treat unlearning data as an opponent to be eliminated, which ignores the potential relationship between the unlearning data and the remaining data.


To better understand federated unlearning and the relationship between unlearning and remaining data, we explored the representation differences between the original model and the model retrained from scratch, as illustrated in \figurename~\ref{Fig:introduction}. Our analysis revealed two key observations within the model's representation space.

First, in the representation space of the model retrained from scratch, the representations of the remaining class data continue to exhibit compact and separable clustering characteristics, similar to those observed in the original model. However, the representations of the unlearning class data become intermingled with those of the remaining classes, indicating a transition from a separable to an inseparable state. This shift ensures that the unlearning model has a diminished generalization capability for the unlearning data while still preserving predictive performance for the remaining data.

Second, the distribution of unlearning class data in the representation space of the model retrained from scratch exhibits a non-random pattern. The representations of the unlearning class data are concentrated within specific representation domains associated with certain classes. This phenomenon reveals a non-random distribution pattern in the forgetting process, suggesting that the unlearning class may be preferentially “forgotten” into particular remaining classes. 

Our observations suggest that unlearning and remaining data are not diametrically opposed but can be transformed within the representation space. While previous unlearning approaches have attempted to eradicate the information of unlearning data by directly targeting influential parameters, our findings indicate that unlearning can be achieved by manipulating parameters guided by the representation behaviors of the model. This insight implies that federated unlearning could be accomplished by transforming the representations of the unlearning class into corresponding remaining class representations.

However, two significant challenges persist: (1) Selecting appropriate transformation classes for the unlearning class. (2) Ensuring consistency in the transformation of the unlearning class across different clients' representation spaces.

When an ideal model retrained from scratch is available, selecting appropriate transformation classes for the unlearning class based on their distribution in the representation space is relatively straightforward. However, in the context of federated unlearning, we lack access to such a model and must instead work with the original model requiring unlearning. Consequently, we must select appropriate transformation classes within the representation space of the original model.

Furthermore, due to the statistical heterogeneity of clients' data, class distributions across different clients may deviate from the global class distribution encompassing all clients. Relying solely on local data distributions can lead to inconsistencies in transforming the unlearning class within the representation space because differences in clients' representation spaces arise from data heterogeneity. Therefore, aligning the transformation of the unlearning class across different clients' representation spaces is crucial.

Based on these insights, we propose Federated Unlearning via Class-aware Representation Transformation (FUCRT), a new method for unlearning data of specific target classes from a trained FL model. FUCRT is built upon two key components: transformation class selection and transformation alignment. In this method, the transformation class selection assigns a transformation class for each unlearning sample based on a global transformation class set and the probability output of an individual sample. These assignments are then utilized for the transformation alignment of unlearning data across different clients. For transformation alignment, we employ dual class-aware contrastive learning, which ensures consistent transformations of unlearning data by optimizing both local and global class-aware representation spaces, while also aligning these transformations across clients. Extensive experimental results demonstrate FUCRT's superior performance in unlearning data of target classes within FL. 

Our main contributions can be summarized as follows:

\begin{itemize}
    \item To the best of our knowledge, this is the first study to explore federated unlearning from a representation perspective. We elucidate the phenomena of federated unlearning within the representation space and present two new observations. Notably, we discover that unlearning and remaining class data are not diametrically opposed but can be transformed within the representation space.

    \item We propose a new method, Federated Unlearning via Class-aware Representation Transformation (FUCRT), designed to facilitate the unlearning of data associated with specific target classes from a trained FL model. By employing a transformation class selection strategy and transformation alignment, FUCRT ensures effective and efficient federated unlearning.

    \item Extensive experimental results validate the effectiveness of our proposed FUCRT, demonstrating its superiority in terms of 100\% erasing guarantee, high model utility preservation, and low retraining rounds. Compared to existing baselines, FUCRT achieves state-of-the-art performance across various datasets under both IID and Non-IID settings.
    
\end{itemize}

\section{Related Works}

\subsection{Federated Learning}

The foundational work in FL is the FedAvg ~\cite{mcmahan2017communication}, which aggregates locally computed model updates to form a global model. This approach addresses privacy concerns and adheres to regulatory requirements by keeping data on-device, thus minimizing the risk of data breaches and preserving user confidentiality. As a result, FL has been widely used in privacy-sensitive fields such as healthcare~\cite{nguyen2022federated,liu2022contribution}, finance~\cite{shao2019stochastic,hahn2019privacy,byrd2020differentially,schreyer2022federated}, and recommendation~\cite{yuan2023federated}. However, FL faces numerous challenges ~\cite{kairouz2021advances,guo2023fedmcsa,guo2022flmjr,guo2022dual,guo2024contribution}, among which the most fundamental is data statistical heterogeneity ~\cite{mendieta2022local,li2020federated,luo2021no}. Data statistical heterogeneity means data distributions among clients are often not independent and identically distributed (Non-IID), which is also a key issue faced in federated unlearning.

\subsection{Machine Unlearning}

The concept of machine unlearning was first introduced by Cao \textit{et al.}~\cite{cao2015towards}, with the aim of developing techniques to more efficiently and accurately erase specific information from the well-trained model, thereby protecting the privacy of data holders~\cite{gupta2021adaptive,tarun2023fast,thudi2022necessity,kurmanji2024towards}. The most straightforward approach is to remove sensitive data and retrain the model from scratch. While this is considered the most effective way to achieve unlearning, its high computational cost has driven researchers to explore alternative solutions. An effective approach is to retrain acceleration. A pioneering approach in this area is SISA, proposed by Bourtoule \textit{et al.}~\cite{bourtoule2021machine}. The core idea is to trade additional storage space for faster unlearning time by dividing the training data into multiple non-overlapping shards and training a submodel for each shard. During the unlearning process, SISA only needs to retrain the submodel containing the specific data sample from a certain checkpoint, significantly reducing the retraining effort. Graves \textit{et al.}~\cite{graves2021amnesiac} achieved retraining acceleration by storing the gradient information for each training batch during the initial training phase. They implemented unlearning by subtracting the gradient updates of specific batches to forget the corresponding data. Another unlearning method involves altering the model using precomputed parameters to adjust the target model~\cite{liu2021federaser,baumhauer2022machine}, which effectively reduces storage overhead compared to retraining acceleration. Additionally, pruning certain gradients from a trained model can also achieve unlearning\cite{schelter2021hedgecut,brophy2021machine,chen2021machine,wu2020deltagrad}. 
Unfortunately, current centralized unlearning methods cannot effectively handle the systems challenges in FL, such as the lack of direct data access and Non-IID data.

\subsection{Federated Unlearning}

Federated Unlearning (FU) aims to enable an FL model to effectively remove the influence of an individual client or eliminate identifiable information associated with a client's partial data, while preserving the integrity of the FL process ~\cite{liu2020federated,su2023asynchronous,liu2023survey}. Based on the specific targets of FU, FU frameworks can be categorized into three distinct levels ~\cite{yang2023survey}: client-level, sample-level, and class-level. Client-level FU seeks to remove the knowledge contributed by specific clients from the global model ~\cite{liu2021federaser}. Sample-level FU aims to eliminate knowledge derived from particular subsets of client data ~\cite{che2023fast}. Class-level FU aims to remove specific classes from the generalization boundary of FL model ~\cite{zhao2023federated,wang2022federated}. In this work, we concentrate on class-level FU to forget data of specific target classes from a trained FL model. 
Yian \textit{et al.} ~\cite{zhao2023federated} proposed the FUMD method, which trains an additional degradation model with the same architecture as the target model on the client requiring unlearning. This degradation model is then integrated with the target model to guide its unlearning process. Based on term frequency-inverse document frequency (TF-IDF), FUDP pruned the most relevant channels associated with the classes to be forgotten, followed by fine-tuning the pruned model for a few rounds ~\cite{wang2022federated}. This method is highly effective when the number of unlearning classes is few. However, as the number of unlearning classes increases, more parts of the model must be pruned, leading to a deterioration in the performance of the remaining classes.

\section{Preliminaries}
\label{sec:preliminaries}

FL is a distributed deep learning paradigm wherein multiple clients, each possessing a local dataset $ \mathcal{D}_{k} = \{ (x_{i}, y_{i}) \}_{i=1}^{|\mathcal{D}_{k}|}$ for the client $k \in [K]$, collaboratively train a global model $\mathcal{M}$ parameterized by $\theta$ without directly sharing their data.

We introduce the notion of FU in the context of supervised classification with Deep Neural Networks (DNNs). Specifically, FU allows for the removal of data samples, denoted as $\mathcal{S}_{k}$ for the $k$-th client, from the trained global model upon request. This unavoidably has an impact on the global model $\mathcal{M}$. In this work, we focus primarily on the case where $\mathcal{S}_{k}$ consists of the samples associated with specific target classes. The goal of FU is to adjust the global model parameters $\theta$ into $\theta^{u}$, such that the performance resembles that of a model $\theta^{*}$ hypothetically trained from scratch using the residual dataset $\mathcal{D}_k \backslash \mathcal{S}_k$. This goal can be theoretically defined as making the distribution of $\mathcal{M}_{\theta^{u}} (\mathcal{D}_k)$ as close to that of $\mathcal{M}_{\theta^{*}} (\mathcal{D}_k \backslash \mathcal{S}_k)$ as possible. Formally, this can be expressed as
\begin{align}
        \mathrm{Pr}(\mathcal{M}_{\theta^{u}} (\mathcal{D}_k) ) \backsimeq \mathrm{Pr}(\mathcal{M}_{\theta^{*}} ( \mathcal{D}_k \backslash \mathcal{S}_k ) ) .
\end{align}
In this work, FU is focused on eliminating the knowledge of specific target classes from the trained global model, under the assumption that both the server and clients are benign. The unlearning process is governed by three principal objectives:

1. Erasing Guarantee: An effective FU solution needs to provide strong guarantees that the data samples of the unlearning classes from clients no longer contribute to the updated model. Specifically, the performance of the unlearning model on the unlearning class data should closely resemble that of a model retrained from scratch without these classes, once the FU process is completed.

2. Model Utility Preservation: Maintaining model performance remains a shared goal for both clients and servers engaged in the FU process. However, it is crucial to acknowledge that unlearning a substantial portion of data samples, which effectively reduces the available training data, inevitably impacts the overall utility of the model. A sophisticated FU method should strive to produce an unlearning model that closely aligns with the performance of a model retrained from scratch, while minimizing the degradation of performance on remaining classes.

3. Efficiency: A naive approach to FU is to retrain the model from scratch on all clients; however, this method is prohibitively time-consuming and computationally expensive, making it impractical. To ensure FU is viable and scalable in real-world applications, it is essential to design an efficient mechanism that minimizes computational overhead while still achieving the desired unlearning outcomes.


\section{Methodology}

\subsection{Overview}

\begin{figure*}[htbp]
  \centering
  \includegraphics[width=0.99\textwidth]{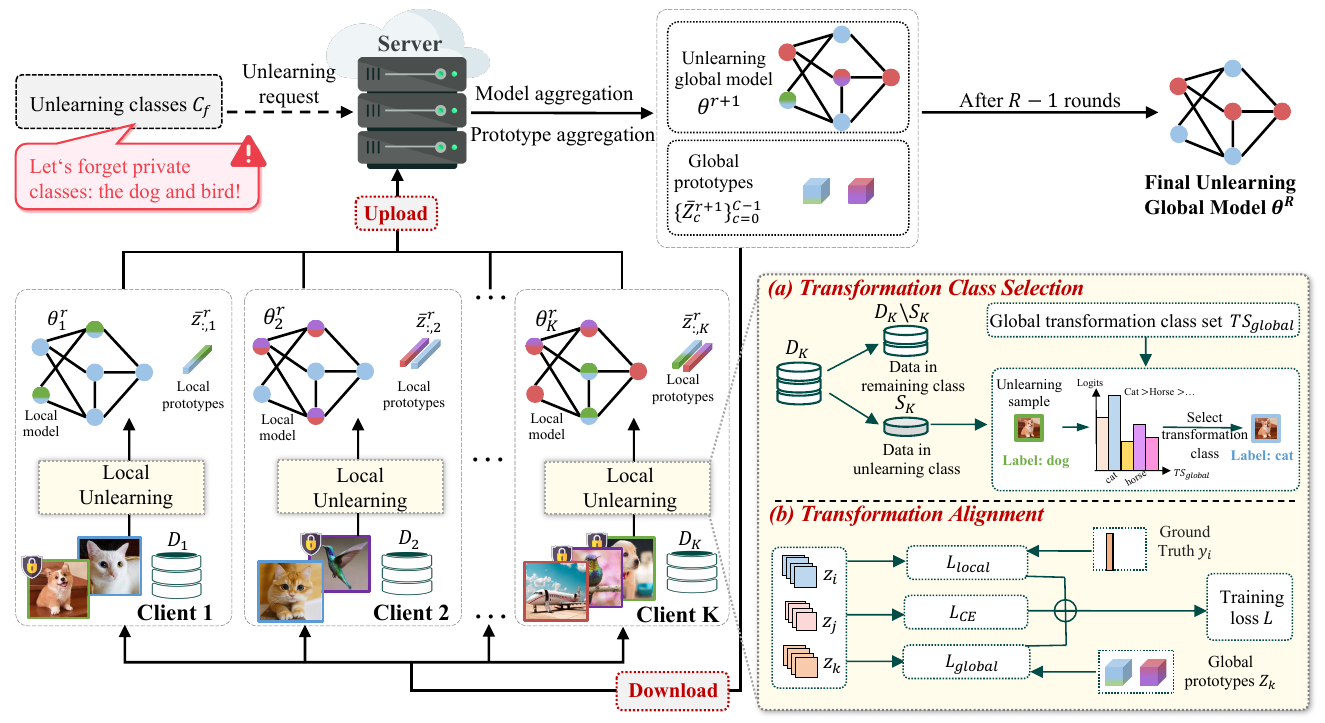}
  \caption{Framework of the proposed FUCRT for unlearning data associated with specific target classes by class-aware representation transformation.}
  \label{fig:framework}
\end{figure*}

As illustrated in \figurename ~\ref{fig:framework}, we present a concise overview of our proposed method, Federated Unlearning via Class-aware Representation Transformation (FUCRT), which is designed to unlearn data of specific target classes. FUCRT maintains the standard client-server architecture of FL while introducing class-aware representation transformation to address FU. We first propose the transformation class selection strategy to identify transformation classes for unlearning samples, as detailed in Section \ref{sec:methodology.tcs}. Then, we introduce transformation alignment to mitigate inconsistency in the transformation process of different clients, as presented in Section \ref{sec:methodology.ta}. Finally, to facilitate a comprehensive understanding of FUCRT workflow, we provide a detailed algorithm in Section \ref{sec:methodology.up}. 

\subsection{Transformation Class Selection}
\label{sec:methodology.tcs}

\subsubsection{Generating the global transformation class set}

Intuitively, in the representation space, the transformation class for an unlearning class is usually the one most likely to be involved in its misclassification. We can use the most probable incorrect prediction for unlearning class data as a reference for selecting the transformation class. However, because the original global model is typically well-trained in FU, there are often few misclassified samples. This scarcity makes relying on the most probable misclassification unreliable.

To leverage the original global model's high performance, we propose using the class with the second-highest probability output for correctly classified samples of the unlearning class as the transformation class. Specifically, we first compute the probability output vector for unlearning class data on each client: $p = M_\theta(x_i), i \in S_k$. For all correctly classified samples, we obtain the second-highest probability outputs and their corresponding classes. We then sum the second-highest probability outputs by class to get $p^{\text{sum}}_c, c \in [C]$, where $C$ is the total number of data categories. The class with the maximum $p^{\text{sum}}_c$ is considered the most likely transformation class, i.e., $\arg\max_{c \in  [C]} \, p^{\text{sum}}_c$.

Considering that an unlearning class may correspond to multiple comparable transformation classes in the representation space, we introduce a threshold $\tau_p$ to determine several most appropriate transformation classes. After sorting $p^{\text{sum}}_c$ in descending order, if $p^{\text{sum}}_j \geq \tau_p \cdot \max_{c \in  [C]} p^{\text{sum}}_c$, the $j$-th class is also confirmed as a transformation class of the unlearning class. This results in a local transformation class set, denoted as $T\!S_k$. $T\!S_k$ is determined by the probability distribution of unlearning class data using the original model on client $k$.

Due to the Non-IID nature of different clients, some may have more samples for certain classes and sparse data for others. To ensure the transformation mappings are well-supported by data, we calculate the local transformation class set only for an unlearning class with sufficient sample data on each client. A client generates its local transformation class set only when its correctly classified sample size for an unlearning class exceeds a threshold $\tau_s$.

After each client computes its local transformation class set $T\!S_k$ based on its data distribution, these sets are uploaded to the server. The server aggregates them to produce a global transformation class set $T\!S_{global}$. This global set is characterized by two factors: the number of elements and the specific class for each element. The number of elements in the global transformed class set, or its size $|T\!S_{global}|$, is determined by analyzing the number of elements from all local transformed class sets and selecting the most frequent count. We then count the elements within all locally uploaded transformed class sets by category, and select the top $|T\!S_{global}|$ categories based on frequency to form the global transformed class set $T\!S_{global}$. The local-to-global aggregation process is formalized as follows:
\begin{equation}
T\!S_{global} = \text{Agg}(T\!S_1, T\!S_2, ..., T\!S_K).
\label{eq:tcs.1}
\end{equation}

\subsubsection{Selecting the transformation class for individual unlearning sample}

After obtaining the global transformation class set, we need to assign a specific transformation class from this set to each unlearning sample. Specifically, during local unlearning training, we examine the model’s output probabilities for each unlearning sample. Based on the probabilities corresponding to the categories in the global transformation class set, we select the category with the highest probability as the transformation class for the current unlearning sample. This strategy leverages both the globally determined transformation class set consensus and the model's current knowledge to select the most appropriate transformation class for each unlearning sample. Formally, we define $t\!s(x_i, y_i)$ as the transformation class for an unlearning sample $(x_i, y_i)$. This assignment can be formalized as follows:
\begin{equation}
t\!s(x_i, y_i) = \arg \max_{c \neq y_i} p_c, c \in T\!S_{global},
\label{eq:tcs.2}
\end{equation}
where $p_c$ represents the probability output for class $c$ on the sample $(x_i, y_i)$, $y_i$ is the original class label of the unlearning sample, $T\!S_{global}$ is the global transformation class set.

This transformation class selection strategy ensures that: (\romannumeral1) The selection respects the global consensus on appropriate transformation classes, maintaining consistency across the FL environment. (\romannumeral2) Each unlearning sample is assigned to the class that is semantically close to its original class, as determined by the model's current understanding.

\begin{algorithm}[h]
	\caption{Federated Unlearning via Class-aware Representation Transformation (FUCRT)}\label{algorithm:fucrt}
	\KwIn{the original global model $\theta^0$, unlearning classes $C_f$, local dataset $\mathcal{D}_k, k \in [K]$, $K$ clients, and maximum unlearning training round $R$.}
	\KwOut{The unlearning global model ${\theta^R}$.}
	\textbf{Procedure} \textit{Server execution}:\\
	\For{$r = 1$ to $R-1$}{
		\For{$k \gets 1$ \textbf{to} $K$ in parallel}{
                send the global model $\theta^{r}$ to the client $k$;\\
			send the global class prototypes $\bar{Z}^{r}=(\bar{Z}_0^{r}, \bar{Z}_1^{r}, ..., \bar{Z}_{C-1}^{r})$to the client $k$;\\
			$\theta^{r}_{k}, \left \{ \bar{z}^{r}_{c,k} \right \}_{c=0}^{C-1} \gets$ \textit{Local Unlearning}$(r,k,\theta^{r}, \bar{Z}^{r})$\\
		}
  
        Obtain unlearning global model $\theta^{r+1}$ by aggregating $r$-th round local models by FedAvg;\\
        
        Obtain global prototype $\bar{Z}^{r+1}_c$ by aggregating local prototypes by Eq.\ref{eq:method.aggproto}; \\
	}
    
    Return the final unlearning global model $\theta^{R}$.\\
    \BlankLine
    \BlankLine
    
	\textbf{Procedure} \textit{Local Unlearning}$(r,k,\theta^{r}, \bar{Z}^{r})$:\\
    Get the local model $\theta^{r}_{k}$ by the global model $\theta^{r}$: $\theta^{r}_{k} \gets \theta^{r}$;\\
    Obtain the new dataset $\mathcal{D}_k^{'}$ by converting unlearning data labels using the transformation class selection strategy: $\mathcal{D}_k^{'} \gets \mathcal{D}_k$;\\
    \For{each batch $\in \mathcal{D}_k^{'}$}{
        compute $\mathcal{L} = \mathcal{L}_{CE} + \lambda_1 \mathcal{L}_{local} + \lambda_2 \mathcal{L}_{global}$ by Eq.\ref{eq:method.losstotal};\\
        $\theta^{r}_k \gets \theta^{r}_k - \eta \nabla \mathcal{L}$;\\

    }
	generate local prototypes $\left \{ \bar{z}^{r}_{c,k} \right \}_{c=0}^{C-1}$ by $\theta^{r}_{k}$ on $\mathcal{D}_k^{'}$;\\
	Return $\theta^{r}_{k}$ and $\left \{ \bar{z}^{r}_{c,k} \right \}_{c=0}^{C-1}$.
\end{algorithm}

\subsection{Transformation Alignment}
\label{sec:methodology.ta}

After obtaining the transformation class results of unlearning data, we convert the labels of the unlearning data to get a new local dataset $\mathcal{D}^{'}_k$ evolved from the old dataset $\mathcal{D}_k$. We can then achieve representation transformation using cross-entropy loss by fine-tuning the original model $\mathcal{M}_{\theta}$ on the new dataset $\mathcal{D}^{'}_k$. Within a batch of $N$ samples, the fundamental cross-entropy loss is expressed as:
\begin{equation}
\label{eq:method.lossce}
    \mathcal{L}_{CE}=-\frac{1}{N}\sum ^{N}_{i=1} {y_{i}log(p)},
\end{equation}
where $p$ is the probability output predicted by the model for the sample $(x_i, y_i)$.

However, the distributed nature of FL often results in heterogeneous data across clients. This statistical heterogeneity engenders discrepancies in the representation spaces of different clients, potentially leading to inconsistencies in the transformation of unlearning classes. Consequently, it is crucial to align the transformation of unlearning data across the representation spaces of different clients.

In the representation space, data can be categorized into unlearning classes and remaining classes, forming a class-aware representation space. The transformation of unlearning data during the unlearning process inevitably leads to a reconstruction of this class-aware representation space. In the context of statistical heterogeneity, relying solely on the local class-aware representation space is insufficient to address the challenges posed by heterogeneity data. On the other hand, focusing exclusively on the global class-aware representation space may result in inconsistencies between the global class-aware representation space and the local representation spaces of individual clients, potentially compromising the unlearning data transformation process during local unlearning training. Therefore, it is essential to consider both the local and global class-aware representation spaces.

To ensure the reliability and consistency of the transformation process, we propose a transformation alignment technique that simultaneously employs local and global class-aware contrastive learning. This technique optimizes the local class-aware representation space through local class-aware contrastive learning while continuously aligning it with the global class-aware representation space via global class-aware contrastive learning. The local class-aware contrastive loss is formulated as:
\begin{equation}
	\mathcal{L}_{local} = -\frac{1}{N}\sum_{i=1}^{N} \frac{1}{N_{y_i}} \sum_{j=1}^{N} \mathbbm{1}_{y_i = y_j} \log \frac{\exp{(z_i \cdot z_j / \tau_t) }}{\sum_{k=1}^{N} \mathbbm{1}_{i \neq k} \exp{(z_i \cdot z_k / \tau_t)}},
    \label{eq:method.losslocal}
\end{equation}
where $z_i$, $z_j$, and $z_k$ indicate the representations of inputs $x_i$, $x_j$, and $x_k$, respectively, extracted by an encoder. For the $i$-th sample, $N_{y_i}$ represents the number of samples in the batch sharing the same label $y_i$. The dot product $z_i \cdot z_j$ quantifies the similarity between representations of the $i$-th and $j$-th samples. The temperature scaling factor $\tau_t$ modulates the distribution's smoothness, while the indicator function $\mathbbm{1}_{condition}$ yields 1 when the condition is true, and 0 otherwise. This loss function $\mathcal{L}_{local}$ aims to enhance intra-class similarity while reducing inter-class similarity, thereby refining the local class-aware representation space during the local unlearning training.

Meanwhile, the global class-aware contrastive loss is defined as:
\begin{equation}
	\mathcal{L}_{global}=-\frac{1}{N} \sum_{i=1}^{N} \sum_{j=0}^{C-1} \mathbbm{1}_{y_i=j} \log \frac{\exp{\left(z_i \cdot \bar{Z}_j / \tau_t\right)}}{\sum_{k=0}^{C-1} \mathbbm{1}_{k \neq j} \exp{\left(z_i \cdot \bar{Z}_k / \tau_t\right)}},
\label{eq:method.lossglobal}
\end{equation}
where $C$ denotes the total number of classes, and $\bar{Z}_j$ represents the global prototype of the $j$-th class, transmitted by the server to facilitate local transformation alignment for samples belonging to class $j$. The global prototype $\bar{Z}_j$ is obtained by aggregating the local prototypes $\bar{z}_{j,k}, k \in [K]$ uploaded by clients on the server, which can be formulated as follows:
\begin{equation}
	\bar{Z}_j = \frac{1}{K}\sum_{k=1}^{K} \bar{z}_{j,k},
\label{eq:method.aggproto}
\end{equation}
where local prototype $\bar{z}_j^k$ is the average representation of class $j$ data on the client $k$, using the model after local unlearning training.

To comprehensively address the challenges of inconsistency in the transformation process, we incorporate both local and global class-aware contrastive losses alongside the conventional cross-entropy loss for local unlearning training. Therefore, the holistic local unlearning training loss function for each client is as follows:
\begin{equation}
\label{eq:method.losstotal}
\mathcal{L} = \mathcal{L}_{CE} + \lambda_1 \mathcal{L}_{local} + \lambda_2 \mathcal{L}_{global},
\end{equation}
where $\lambda_1$ and $\lambda_2$ are coefficients governing the contributions of local and global contrastive losses, respectively. Through this dual class-aware contrastive learning, we simultaneously optimize the local and global class-aware representation spaces while continuously aligning them. This process ensures the reliability and consistency of unlearning class data's representation transformation in the representation space, effectively addressing the challenge of inconsistent transformation process posed by Non-IID data in FU.

\subsection{Unlearning Process}
\label{sec:methodology.up}

The unlearning process of FUCRT begins with the original global model $\theta^{0}$ and unlearning classes $C_f$ that are requested to be forgotten. We employ a transformation class selection strategy to select a specific transformation class for each sample from unlearning data. Next, during transformation alignment, we propose local and global class-aware contrastive losses for unlearning training. These losses ensure consistent transformations of unlearning data by optimizing both local and global class-aware representation spaces and aligning these transformations across clients. Additionally, there are cases where clients only possess data from the remaining classes. In these situations, the client does not need to perform representation transformation but only needs to execute transformation alignment, optimizing the local class-aware representation space to align with the global space. After learning for $R-1$ rounds, FUCRT outputs the final unlearning global model $\theta^{R}$, which has effectively forgotten the data from the unlearning classes. Algorithm \ref{algorithm:fucrt} presents the complete unlearning process of FUCRT, detailing both the server execution and the procedure of local unlearning on clients.

\section{Experimental Evaluation}

\subsection{Experimental Setup}
{\bf Datasets and Network Architecture.}
We conduct extensive experiments on the following four real-world datasets: 

\begin{itemize}
\item \textbf{CIFAR10}~\cite{krizhevsky2009learning} is a widely used benchmark dataset for image classification tasks. CIFAR10 consists of 60,000 32x32 color images across 10 classes, with 6,000 images per class. The data of each class contains 5,000 training images and 1,000 testing images.

\item\textbf{CIFAR100}~\cite{krizhevsky2009learning} has 60,000 32x32 color images and contains 100 classes with 600 images for each class. This dataset presents a more challenging classification problem due to its finer-grained categories. The data of each class contains 500 training and 100 test images.

\item\textbf{FMNIST}~\cite{xiao2017fashion} comprises 70,000 28x28 grayscale images divided into 10 fashion categories, with 7,000 images per class. This dataset contains a training set of 60,000 examples and a test set of 10,000 examples.

\item\textbf{EuroSAT}~\cite{helber2019eurosat} is a publicly available georeferenced image dataset constructed from satellite data. It contains 27,000 64x64 color satellite images from various regions in Europe. The dataset was split into a training set and a test set in an 80\% : 20\% ratio.
\end{itemize}

For the CIFAR10, FMNIST, and EuroSAT datasets, we employ the widely used ResNet18 as the default network architecture \cite{he2016deep}. For CIFAR100, we opt for the more powerful ResNet50 as the network architecture.

\begin{table*}[ht]

    \caption{ Accuracy and F1 on CIFAR10, CIFAR100, FMNIST, and EuroSAT datasets under the IID and Non-IID settings (\%). The unlearning class percentage is 10\%.}

    \centering

        \resizebox{0.895\textwidth}{!}{

    \begin{tabular}{c|c|cccc|cccc}

    \toprule

    \multirow{3}{*}{\textbf{Dataset}}   & \multirow{3}{*}{\textbf{Method}}       & \multicolumn{4}{c|}{\textbf{IID}}                                                 & \multicolumn{4}{c}{\textbf{Non-IID}}                                             \\ \cline{3-10}

                                        &                                        & \multicolumn{2}{c}{\textbf{Unlearning}} & \multicolumn{2}{c|}{\textbf{Remaining}} & \multicolumn{2}{c}{\textbf{Unlearning}} & \multicolumn{2}{c}{\textbf{Remaining}} \\ \cline{3-10}

                                        &                                        & ACC                          & F1       & ACC              & F1                   & ACC                & F1                 & ACC               & F1                 \\ \midrule

    \multirow{8}{*}{\textbf{CIFAR10}}  & Origin                                 & 93.56                        & 96.50    & 87.70            & 87.63                & 91.77              & 95.34              & 86.14             & 86.01              \\

                                        & From-scratch                           & 0.00                         & 0.00     & 87.99            & 87.65                & 0.00               & 0.00               & 85.44             & 85.00              \\

                                        & Fine-tune                              & 52.04                        & 67.28    & 89.82            & 89.63                & 17.16              & 31.75              & 90.08             & 89.74              \\

                                        & Gradient-ascent~\cite{wu2022federated} & 0.00                         & 0.00     & 10.94            & 16.44                & 0.00               & 0.00              & 11.15             & 19.91              \\

                                        & FUDP~\cite{wang2022federated}          & 0.00                         & 0.00     & 84.85            & 84.30                & 0.00               & 0.00               & 85.68             & 85.08              \\

                                        & FUMD~\cite{zhao2023federated}          & 0.00                         & 0.00     & 82.38            & 81.89                & 0.00               & 0.00               & 82.60             & 82.02              \\

                                        & \textbf{FUCRT} (Ours)                  & 0.00                         & 0.00     & 90.06            & 89.83                & 0.00               & 0.00               & 89.82             & 89.55              \\ \midrule

    \multirow{8}{*}{\textbf{CIFAR100}} & Origin                                 & 75.48                        & 90.81    & 75.88            & 82.65                & 75.54              & 90.71              & 75.65             & 82.50              \\

                                        & From-scratch                           & 0.00                         & 0.00     & 76.73            & 86.04                & 0.00               & 0.00               & 77.00             & 86.44              \\

                                        & Fine-tune                              & 49.33                        & 84.99    & 76.38            & 86.29                & 48.98              & 83.24              & 76.59             & 86.40              \\

                                        & Gradient-ascent~\cite{wu2022federated} & 0.00                         & 0.00     & 5.08             & 24.03                & 0.00               & 0.00               & 3.01              & 14.12              \\

                                        & FUDP~\cite{wang2022federated}          & 0.03                         & 17.50    & 72.25            & 78.77                & 8.88               & 68.28              & 69.71             & 76.71              \\

                                        & FUMD~\cite{zhao2023federated}          & 0.00                         & 0.00     & 8.45             & 44.98                & 0.00               & 0.00               & 8.95              & 44.75              \\

                                        & \textbf{FUCRT} (Ours)                  & 0.00                         & 0.00     & 75.25            & 85.70                & 0.00               & 0.00               & 74.88             & 85.43              \\ \midrule

    \multirow{8}{*}{\textbf{FMNIST}}    & Origin                                 & 98.46                        & 99.19    & 89.21            & 88.92                & 98.87              & 99.41              & 86.59             & 86.08              \\

                                        & From-scratch                           & 0.00                         & 0.00     & 91.35            & 91.09                & 0.00               & 0.00               & 87.54             & 86.89              \\

                                        & Fine-tune                              & 91.79                        & 95.55    & 91.34            & 91.02                & 30.47              & 46.89              & 91.19             & 90.74              \\

                                        & Gradient-ascent~\cite{wu2022federated} & 0.00                         & 0.00     & 7.92             & 18.83                & 0.00               & 0.00               & 8.80              & 17.41              \\

                                        & FUDP~\cite{wang2022federated}          & 0.00                         & 0.00     & 89.31            & 89.05                & 0.00               & 0.00               & 85.17             & 84.41              \\

                                        & FUMD~\cite{zhao2023federated}          & 0.00                         & 0.00     & 88.41            & 88.16                & 0.00               & 0.00               & 88.86             & 88.34              \\

                                        & \textbf{FUCRT} (Ours)                  & 0.00                         & 0.00     & 90.82            & 90.41                & 0.00               & 0.00               & 90.95             & 90.53              \\ \midrule

    \multirow{8}{*}{\textbf{EuroSAT}}   & Origin                                 & 94.09                        & 96.67    & 94.70             & 96.75                & 87.55              & 93.07              & 84.40             & 90.32              \\

                                        & From-scratch                           & 0.00                         & 0.00     & 94.94            & 97.04                & 0.00               & 0.00               & 86.69             & 91.43              \\

                                        & Fine-tune                              & 1.55                         & 4.33     & 96.33            & 97.87                & 0.11               & 1.16               & 92.58             & 95.65              \\

                                        & Gradient-ascent~\cite{wu2022federated} & 0.00                         & 0.00     & 12.31            & 41.11                & 0.00               & 0.00               & 10.02             & 92.33              \\

                                        & FUDP~\cite{wang2022federated}          & 0.00                         & 0.00     & 90.47            & 94.23                & 0.00               & 0.00               & 81.00             & 88.44              \\

                                        & FUMD~\cite{zhao2023federated}          & 0.00                         & 0.00     & 89.90            & 94.24                & 0.00               & 0.00               & 83.01             & 89.72              \\

                                        & \textbf{FUCRT} (Ours)                  & 0.00                         & 0.00     & 94.92            & 97.04                & 0.00               & 0.00               & 87.38             & 92.89              \\ \bottomrule

    \end{tabular}}

    \label{table:exp1}

    \end{table*}


{\bf Baselines.}
We adopt five most relevant federated unlearning algorithms as our baselines:

\begin{itemize}
\item
\textbf{From-scratch}:
The federated learning model is retrained from scratch using only the remaining data, producing an ideal unlearning model.

\item\textbf{Fine-tune}:
The original global model is fine-tuned using the remaining data to obtain the unlearning model.
\item\textbf{Gradient-ascent}~\cite{wu2022federated}:
This method applies reverse stochastic gradient ascent to tighten the model’s generalization boundary, effectively removing its capacity to classify certain data.
\item\textbf{FUDP}~\cite{wang2022federated}:
Utilizing channel pruning based on term frequency-inverse document frequency (TF-IDF), this method prunes channels most correlated with the unlearning class data, followed by fine-tuning with the remaining class data.
\item\textbf{FUMD}~\cite{zhao2023federated}:
This method implements a knowledge erasure strategy called momentum degradation and provides a memory guidance mechanism to alleviate the performance decline caused by the erasure.
\end{itemize}
{\bf Federated Learning Setting and Details.}
In our experiments, we set the total number of clients at 20, with all clients selected per round. The default Dirichlet coefficient $\delta$=0.5 for the Non-IID scenario. We use SGD \cite{DBLP:journals/ijon/Amari93} as the optimizer during local training on each client, running one local epoch per round. The batch and the prototype size are set to 256 and 512. The learning rate is fixed at 0.01 for both the original model training and unlearning (except for CIFAR100, where the original model was trained using a cosine-annealed learning rate starting at 0.1). 

\subsection{Erasing Guarantee and Model Utility Preservation}
\label{sec:experimental.egmup}

An effective unlearning algorithm must ensure that the unlearning model exhibits poor generalization on unlearning classes while preserving performance on remaining classes. To evaluate the effectiveness of the unlearning model, we utilize two primary metrics: Accuracy (F1) on unlearning classes and Accuracy (F1) on remaining classes.

First, we present a comparison of the performance of baselines and FUCRT on four datasets, each with 10\% of the classes unlearned, as shown in \tablename ~\ref{table:exp1}. Overall, all unlearning methods reduce the influence of unlearning classes to varying degrees. Fine-tune method can indeed preserve the generalization performance on remaining classes by using a large amount of training on the remaining classes. However, as an unlearning method, it fails to guarantee the effectiveness of erasure. For instance, on FMNIST, the accuracy of the unlearning classes reached 91.97\%, almost identical to that of the remaining classes. Even on EuroSAT, the best-performing dataset, the unlearning class accuracy remained around 1\%. In contrast, Gradient-ascent method, which disrupts gradients using only the unlearning class data, cannot preserve the performance of the remaining classes. Consequently, the highest test accuracy of remaining classes across the four datasets was only 12.31\%.

\begin{figure*}[htbp]
  \centering
  \includegraphics[width=1.0\textwidth]{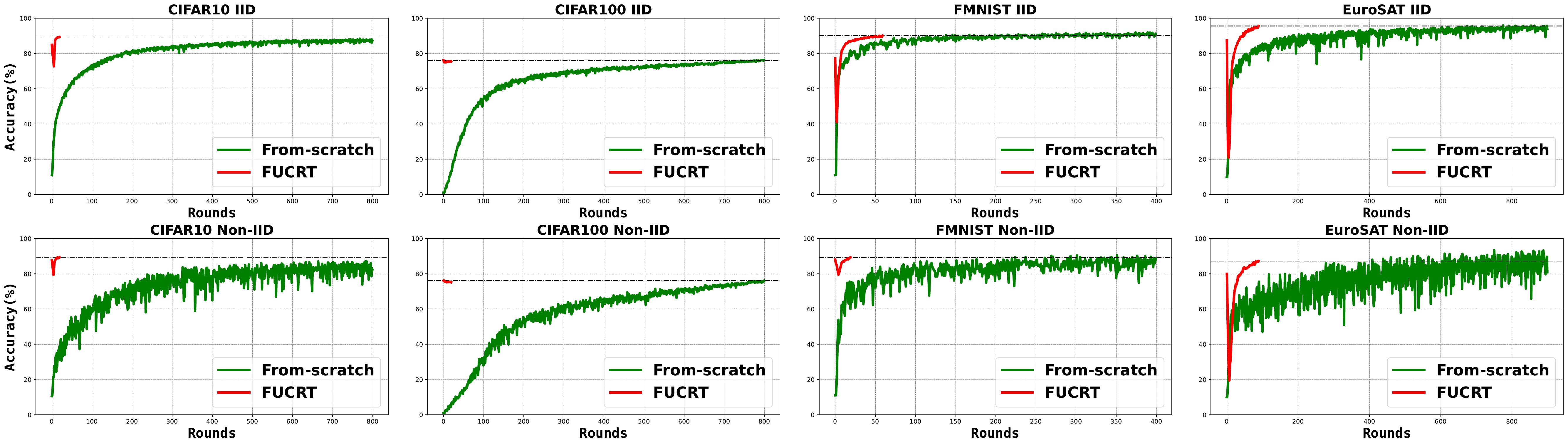}
  \caption{Test accuracy of remaining class data over training rounds. (Federated unlearning the 10\% categories of CIFAR10, CIFAR100, FMNIST, and EuroSAT datasets from pre-trained models.)}
  \label{fig:efficiency}
\end{figure*}

Additionally, we analyzed two state-of-the-art (SOTA) methods in class-level federated unlearning scenarios. FUDP, which employs channel pruning, generally proves effective for forgetting the unlearning classes (with 0.00\% test accuracy on CIFAR10, FMNIST, and EuroSAT datasets and up to 8.88\% on CIFAR100 dataset). However, this method does not maintain consistent effectiveness across all datasets for remaining classes. This is primarily due to the need to prune more model parameters when there is less differentiation between classes, which negatively impacts the model's performance on remaining classes. Even after fine-tuning, accuracy on remaining classes rarely reaches the pre-unlearning level, with a drop of 5.94\% on CIFAR100 Non-IID data, and an average decrease of 2.81\% across four datasets under Non-IID settings. Another recent unlearning method, FUMD, requires clients to retain a degradation model trained solely on the remaining classes to guide the unlearning of the target model. While it generally achieves effective unlearning, the target model struggles to generalize well on remaining classes due to the degradation model's focus solely on the unlearning classes. On CIFAR10 with the Non-IID setting, the remaining class accuracy dropped by 3.54\%. Notably, on CIFAR100, the unlearning class accuracy was below 9\%, mainly because training CIFAR100 on a ResNet50 degradation model requires more convergence rounds than training other datasets on a ResNet18 model.
As illustrated in \tablename ~\ref{table:exp1}, FUCRT guarantees effective unlearning on the four datasets. Across all datasets, our method completely reduces the accuracy on unlearning classes to 0.00\%, while also improving the performance on remaining classes, often surpassing the original global model. Notably, under Non-IID settings, the average improvement is 2.56\%, higher than the average improvement under IID settings (0.89\%). This demonstrates the advantage of our method in Non-IID scenarios, as FUCRT aligns class representations across clients, reducing performance loss due to representation differences. Furthermore, in experiments with CIFAR10, FMNIST, and EuroSAT datasets, our method often slightly improves the remaining class accuracy compared to both the original global model and the model retrained from scratch. For instance, on CIFAR10 under the Non-IID setting, the accuracy on remaining classes increased by 3.68\% compared to the original model and by 4.37\% compared to the model retrained from scratch. This is attributed to our method’s use of global class prototypes to optimize the representation space, further enhancing the generalization on remaining classes. This utility will also be directly demonstrated in subsequent analyses. However, we note that this improvement is not observed in datasets like CIFAR100 with a large number of classes, as the representation space in CIFAR100 is more robust, and a few unlearning rounds are insufficient to properly reorganize such a large representation space. Ultimately, FUCRT achieves performance levels comparable to the retraining from scratch method in terms of both unlearning and remaining classes, surpassing other baselines.

\subsection{Efficiency}

To evaluate the efficiency of our method, we compared the unlearning training process of FUCRT to retraining from scratch. \figurename ~\ref{fig:efficiency} illustrates the test accuracy curves of remaining class data over training rounds. 

On the CIFAR10, FMNIST, and EuroSAT datasets, we observed a brief performance drop for remaining classes during the initial unlearning stage. This temporary decline is attributed to the class-aware representation transformation disrupting the original representation space. However, FUCRT quickly rebuilds the new representation space, restoring performance on remaining classes to pre-unlearning levels.

Interestingly, this recovery process was not observed on the CIFAR100 dataset. As previously analyzed, the robustness of the CIFAR100 representation space means that a limited number of unlearning rounds are insufficient to significantly reconstruct its representation space.

FUCRT demonstrates superior efficiency compared to retraining from scratch, requiring fewer training rounds while achieving better performance on remaining classes. Specifically, FUCRT is 40, 40, 13.3, and 10.2 times more efficient than the From-scratch method on CIFAR10, CIFAR100, FMNIST, and EuroSAT, respectively.

\subsection{Ablation Studies}

\begin{table*}[htbp]
\centering
\caption{Ablation studies on CIFAR10, CIFAR100, FMNIST, and EuroSAT datasets. The unlearning class percentage is 10\%.}
\resizebox{\textwidth}{!}{
\begin{tabular}{c|cccc|cccc|cccc|cccc}\toprule
             \multirow{3}{*}{\textbf{Method}} & \multicolumn{4}{c|}{\textbf{CIFAR10}}                           & \multicolumn{4}{c|}{\textbf{CIFAR100}}                           & \multicolumn{4}{c|}{\textbf{FMNIST}}                            & \multicolumn{4}{c}{\textbf{EuroSAT}} \\\cline{2-17}
             & \multicolumn{2}{c}{\textbf{IID}} & \multicolumn{2}{c|}{\textbf{Non-IID}} & \multicolumn{2}{c}{\textbf{IID}} & \multicolumn{2}{c|}{\textbf{Non-IID}} & \multicolumn{2}{c}{\textbf{IID}} & \multicolumn{2}{c|}{\textbf{Non-IID}} & \multicolumn{2}{c}{\textbf{IID}} & \multicolumn{2}{c}{\textbf{Non-IID}} \\ \cline{2-17}
             & ACC       & F1    & ACC        & F1     & ACC       & F1    & ACC         & F1      & ACC       & F1    & ACC         & F1      & ACC       & F1    & ACC         & F1      \\ \midrule
\textbf{FUCRT}    & \textbf{90.06}     & \textbf{89.83}       & \textbf{89.82}       & \textbf{89.55}         & \textbf{75.25}     & \textbf{85.70 }      & \textbf{74.88}       & \textbf{85.43}         & \textbf{90.82}     &\textbf{ 90.41 }      & \textbf{90.95}       & \textbf{90.53}         & \textbf{94.92}     & \textbf{97.04}       &\textbf{ 87.38 }      & \textbf{92.89}         \\ \midrule
w/o TCS    & 89.60     & 89.43       & 89.24       & 88.97         & 74.41     & 81.81       & 74.62       & 81.80         & 88.81     & 88.34       & 86.20       & 85.22         & 90.19     & 93.21       & 84.82       & 90.51         \\
$\Delta$            & 0.45 $\downarrow$      & 0.40 $\downarrow$       & 0.58$\downarrow$        & 0.58$\downarrow$          & 0.85$\downarrow$     & 3.89$\downarrow$        & 0.26$\downarrow$       & 3.63$\downarrow$          & 2.00$\downarrow$      & 2.07$\downarrow$        & 4.75$\downarrow$        & 5.31$\downarrow$          & 4.74$\downarrow$      & 3.83$\downarrow$        & 2.56$\downarrow$        & 2.38$\downarrow$          \\\midrule
w/o $\mathcal{L}_{local}$  & 89.39     & 89.17       & 20.87       & 57.87         & 75.00     & 81.54       & 74.29       & 81.59         & 21.54     & 59.00       & 21.39       & 58.64         & 23.10     & 96.54       & 21.39       & 79.64         \\
$\Delta$             & 0.67$\downarrow$      & 0.66$\downarrow$        & 68.94$\downarrow$       & 31.68$\downarrow$         & 0.25$\downarrow$     & 4.16$\downarrow$        & 0.59$\downarrow$       & 3.84$\downarrow$          & 69.27$\downarrow$     & 31.41$\downarrow$       & 69.56$\downarrow$       & 31.89$\downarrow$         & 71.82$\downarrow$     & 0.50$\downarrow$        & 66.00$\downarrow$       & 13.26$\downarrow$         \\\midrule
w/o $\mathcal{L}_{global}$ & 89.97     & 89.73       & 89.76       & 89.51         & 75.19     & 81.35       & 74.83       & 81.23         & 89.81     & 89.31       & 89.84       & 89.50         & 90.50     & 93.07       & 83.41       & 89.39         \\
$\Delta$             & 0.08$\downarrow$      & 0.10$\downarrow$        & 0.05$\downarrow$        & 0.05$\downarrow$          & 0.06$\downarrow$     & 4.35$\downarrow$        & 0.06$\downarrow$        & 4.20$\downarrow$          & 1.01$\downarrow$      & 1.09$\downarrow$        & 1.11$\downarrow$        & 1.03$\downarrow$          & 4.42$\downarrow$      & 3.97$\downarrow$        & 3.97$\downarrow$        & 3.51$\downarrow$    \\ \bottomrule     
\end{tabular}}
\label{table:ablation}
\end{table*}

\begin{figure}[htbp]
  \centering
  \subfloat[FUCRT]{
    \includegraphics[width=0.24\textwidth]{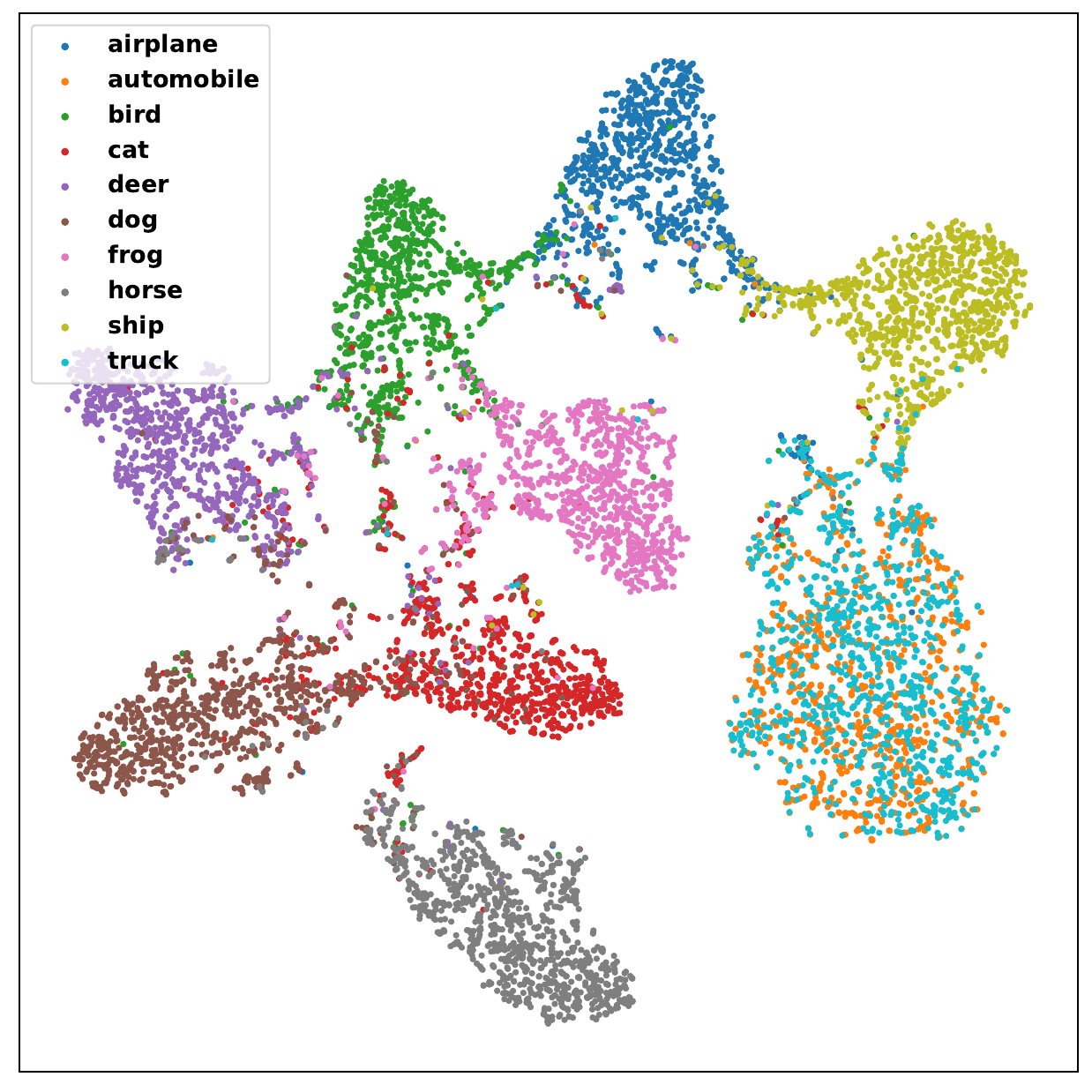}
  }
  \subfloat[FUCRT-random]{
    \includegraphics[width=0.24\textwidth]{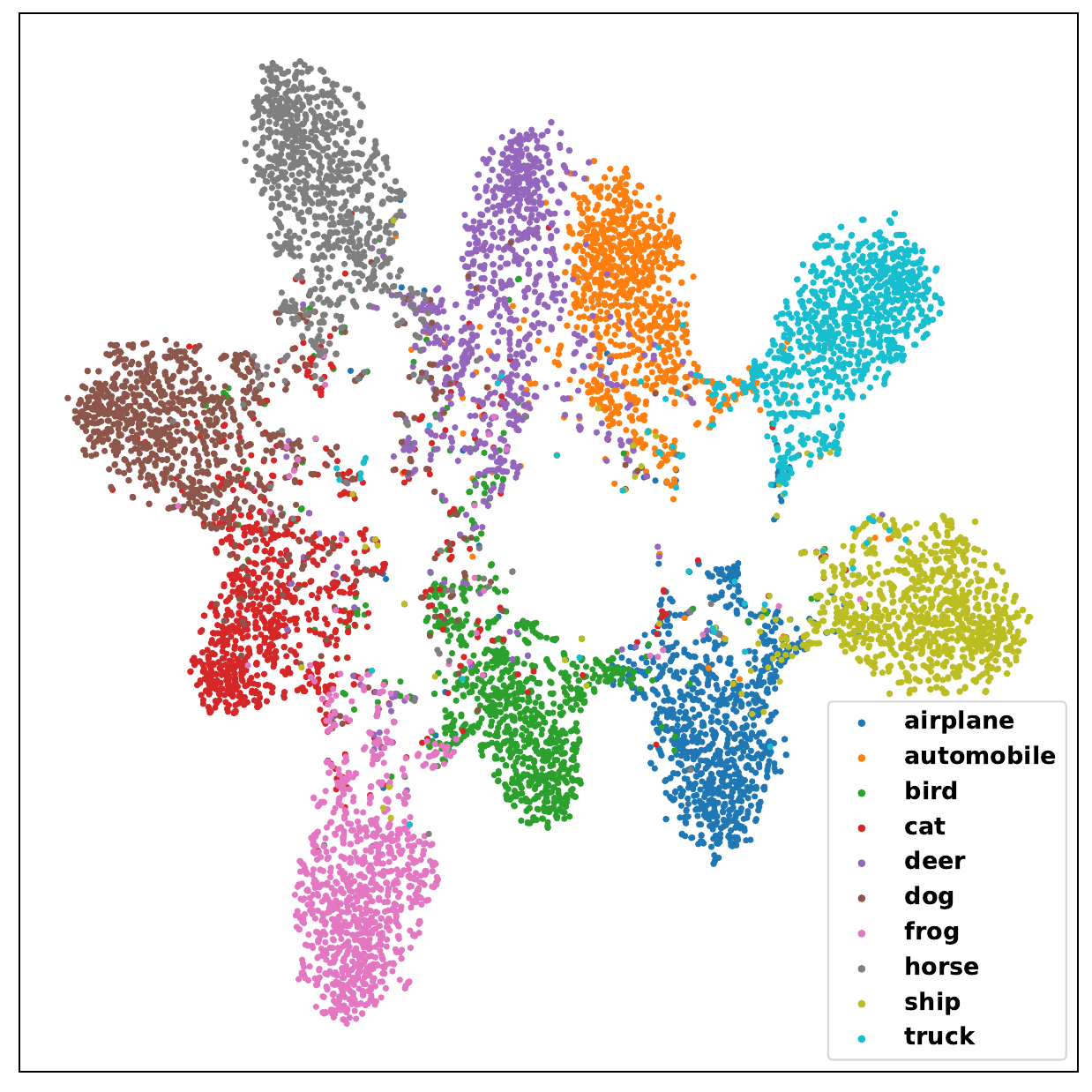}
  }
  \caption{ Comparison between TCS and a random label strategies. (The random label strategy maps an unlearning class to a random class. The unlearning class is ‘automobile’.)}
  \label{fig:ablation}
\end{figure}

To better understand FUCRT, we conducted ablation studies to evaluate the impact of its key components. We studied three key components: transformation class selection strategy (TCS), local class-aware contrastive loss ($\mathcal{L}_{local}$), and global class-aware contrastive loss ($\mathcal{L}_{global}$), as shown in \tablename ~\ref{table:ablation}.

Specifically, to verify the necessity of the TCS strategy, we replaced the TCS in FUCRT with a more naive approach, a random label strategy, which maps unlearning classes to random classes. As illustrated in \figurename ~\ref{fig:ablation}, the unlearning class failed to align with a random class when using the random approach. This resulted in two distinct clusters in the representation space, contrasting with a fused cluster observed in FUCRT. Furthermore, \tablename ~\ref{table:ablation} shows that the random label strategy led to a decline in accuracy and F1 for remaining classes across various datasets. These findings underscore the importance of the TCS strategy in guiding FUCRT toward optimal forgetting directions. 

In addition, to evaluate the effectiveness of transformation alignment, we conducted experiments by removing either the local or global contrastive loss from FUCRT. Our results reveal that relying solely on global prototypes to align inter-client representation differences is insufficient for ensuring effective local unlearning. The accuracy difference for remaining classes reached up to 71.82\% when omitting the local contrastive loss. 
Meanwhile, due to the differences in representation space caused by differences in data distribution of different clients, the lack of global contrastive loss will also affect the transformation alignment process, resulting in a performance decrease of the remaining classes.

The ablation studies collectively demonstrate that each component of FUCRT contributes to its overall performance. The TCS strategy provides optimal forgetting directions, while the combination of local and global contrastive losses ensures effective alignment both within and across clients. 

\subsection{Privacy Guarantee}

\begin{figure}[tbp]
  \centering
  \subfloat[Unlearning 10\% classes]{
    \includegraphics[width=0.24\textwidth]{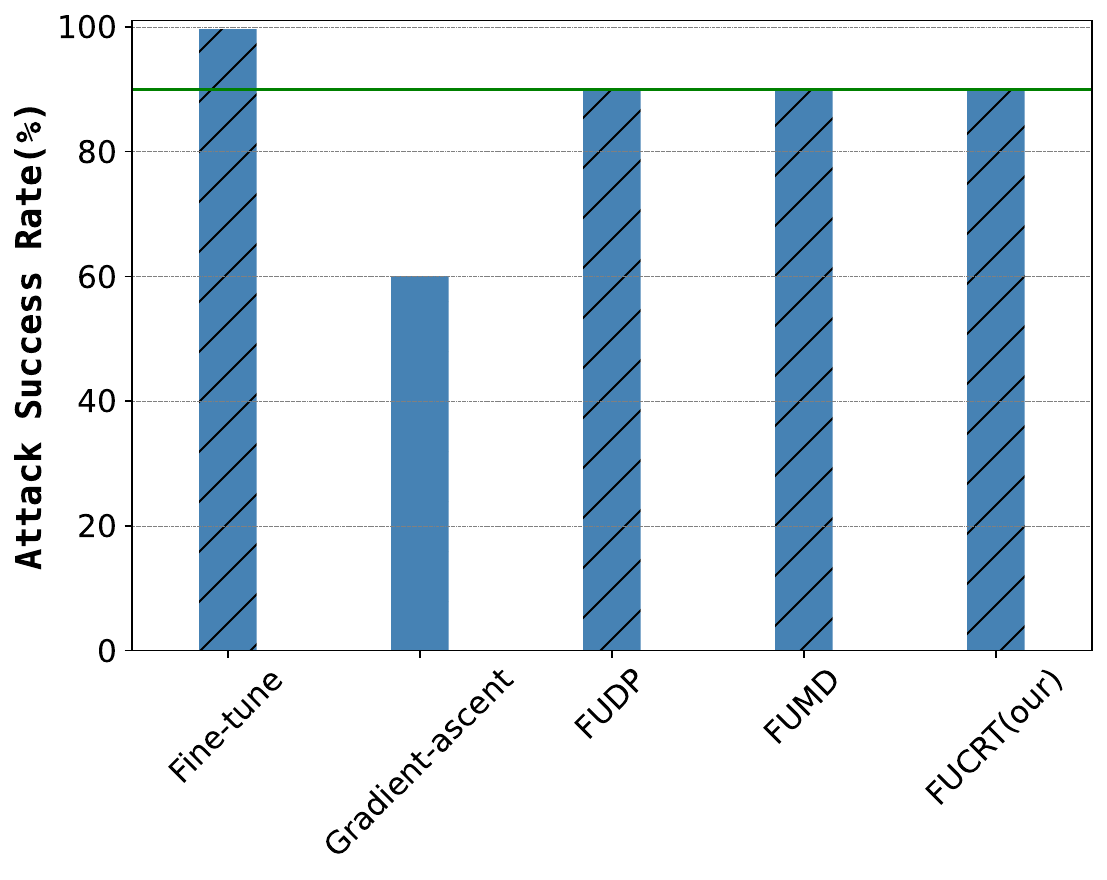}
  }
  \subfloat[Unlearning 30\% classes]{
    \includegraphics[width=0.24\textwidth]{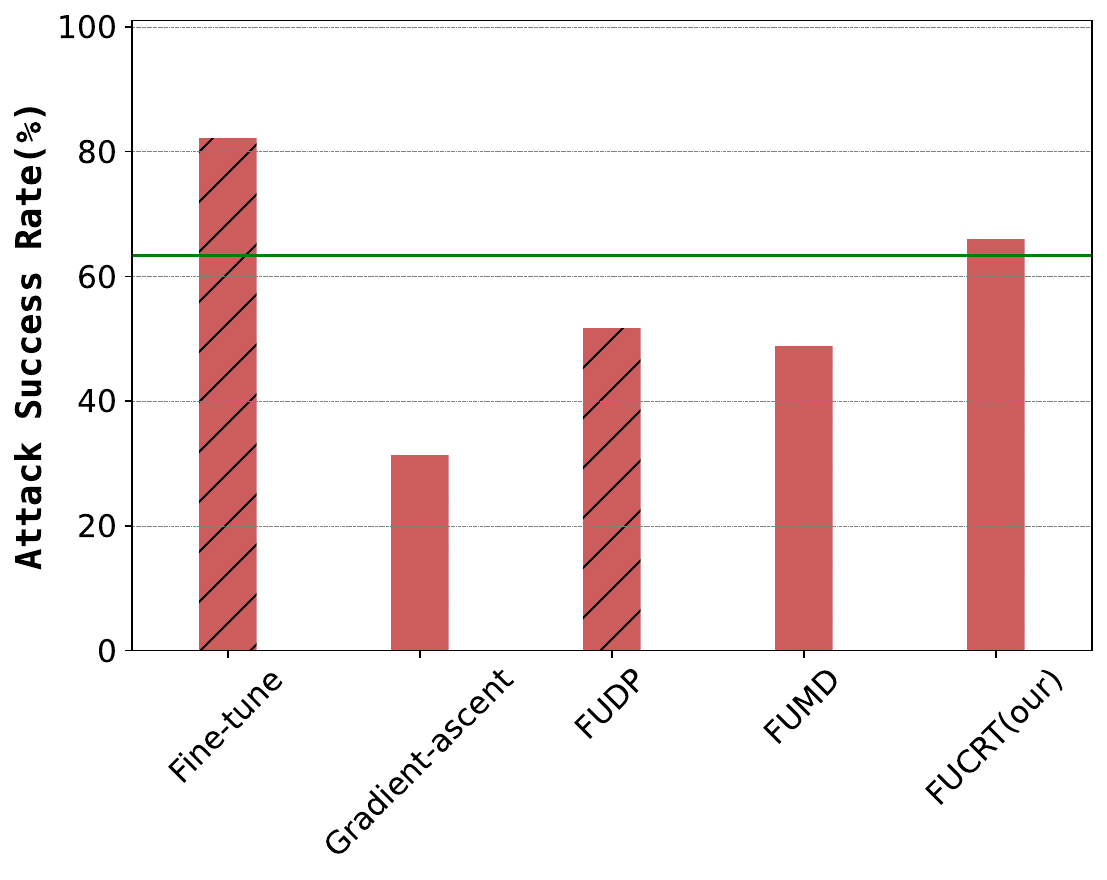}
  }
  \caption{Privacy guarantee of different unlearning methods. (The green line denotes the ASR of retraining from scratch.)}
  \label{fig:mia}
\end{figure}

To assess the privacy guarantees of FUCRT, we conducted a comparative analysis of privacy leakage for the unlearning classes between the baselines and FUCRT. Specifically, we trained Membership Inference Attack (MIA) models on CIFAR10 with the setting of unlearning 10\% and 30\% classes. We attacked the unlearning models resulting from all unlearning methods and plotted bar charts to compare their Attack Success Rates (ASRs), as shown in \figurename ~\ref{fig:mia}. The green line denotes the ASR of retraining from scratch method (From-scratch). An unlearning method with an ASR closer to this green line indicates less leakage of private information on unlearning classes. An ASR approaching 100\% suggests that information from the unlearning classes remains in the model, whereas an ASR approaching 0\% implies excessive disruption to the information of the remaining classes.

The results indicate that, for both unlearning proportions, the Fine-tune method retained too much information from the unlearning classes, as it was unable to effectively remove their influence from the model. This observation is also evident in the t-SNE plots presented in the next subsection. Conversely, the Gradient-ascent method excessively disrupted the representation space, leading to significant loss of information about the remaining classes.

Furthermore, while FUDP and FUMD achieved ASRs close to that of retraining from scratch when only 10\% of classes was forgotten, their ASRs deviated from the desired level when 30\% of classes was forgotten. In the case of FUDP, this was primarily due to the need to prune more parts of the model as more classes were unlearned, which adversely affected the information about the remaining classes. Similarly, for FUMD, the ASR decreased with an increasing number of unlearned classes. In contrast, FUCRT achieved ASRs close to those of retraining from scratch in both the 10\% and 30\% unlearning scenarios, indicating that FUCRT provides superior privacy guarantees compared to other baselines.

\subsection{Representation Space Analysis}
\label{sec:experimental.rsa}

\begin{figure*}[htbp]
  \centering
  \includegraphics[width=1.0\textwidth]{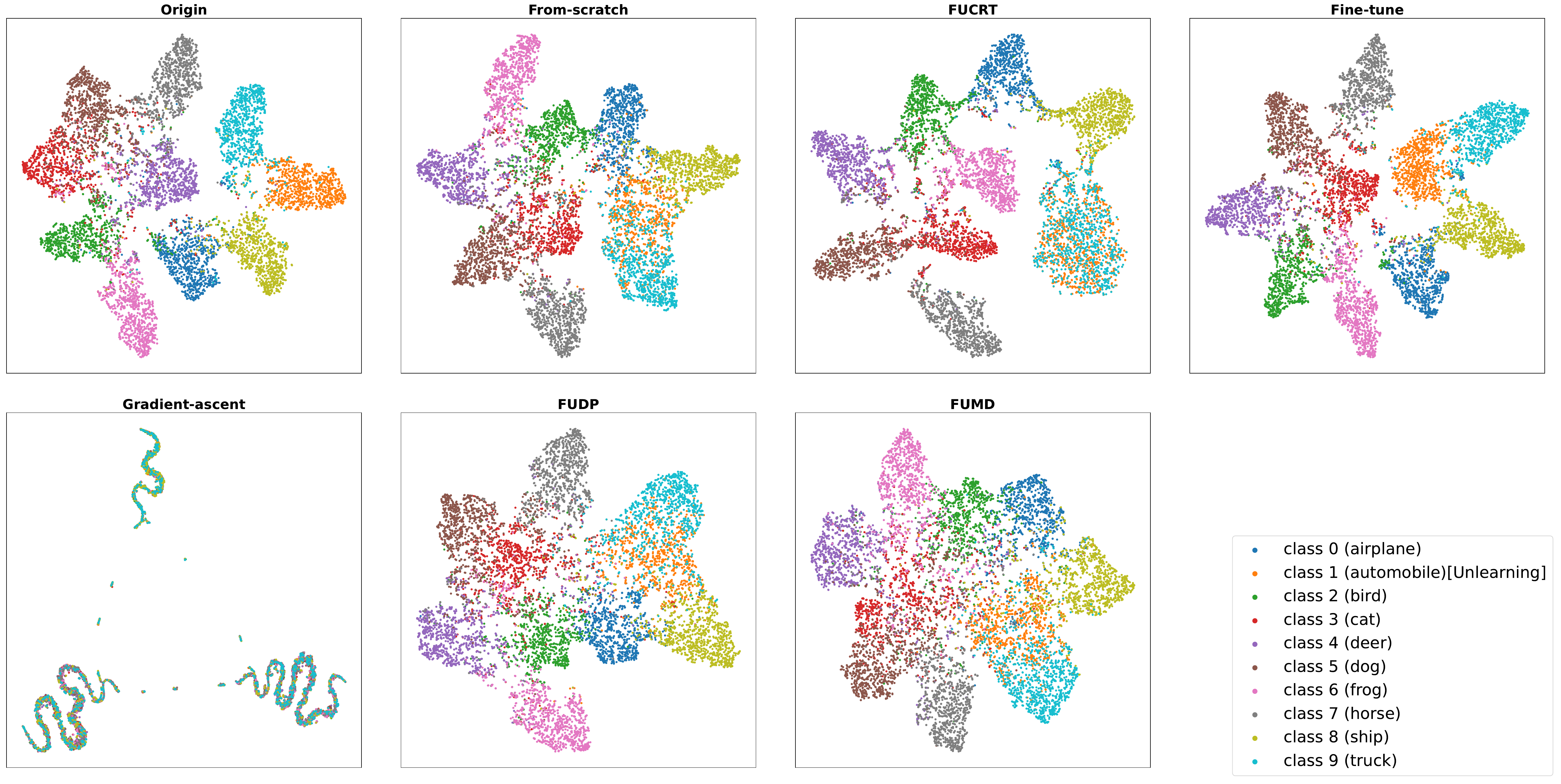}
  \caption{t-SNE visualization of the representation space distribution of unlearning methods with 10\% of the classes forgotten on CIFAR10 with the IID setting.}
  \label{Fig:tsne1}
\end{figure*}

\begin{figure*}[htbp]
  \centering
  \includegraphics[width=1.0\textwidth]{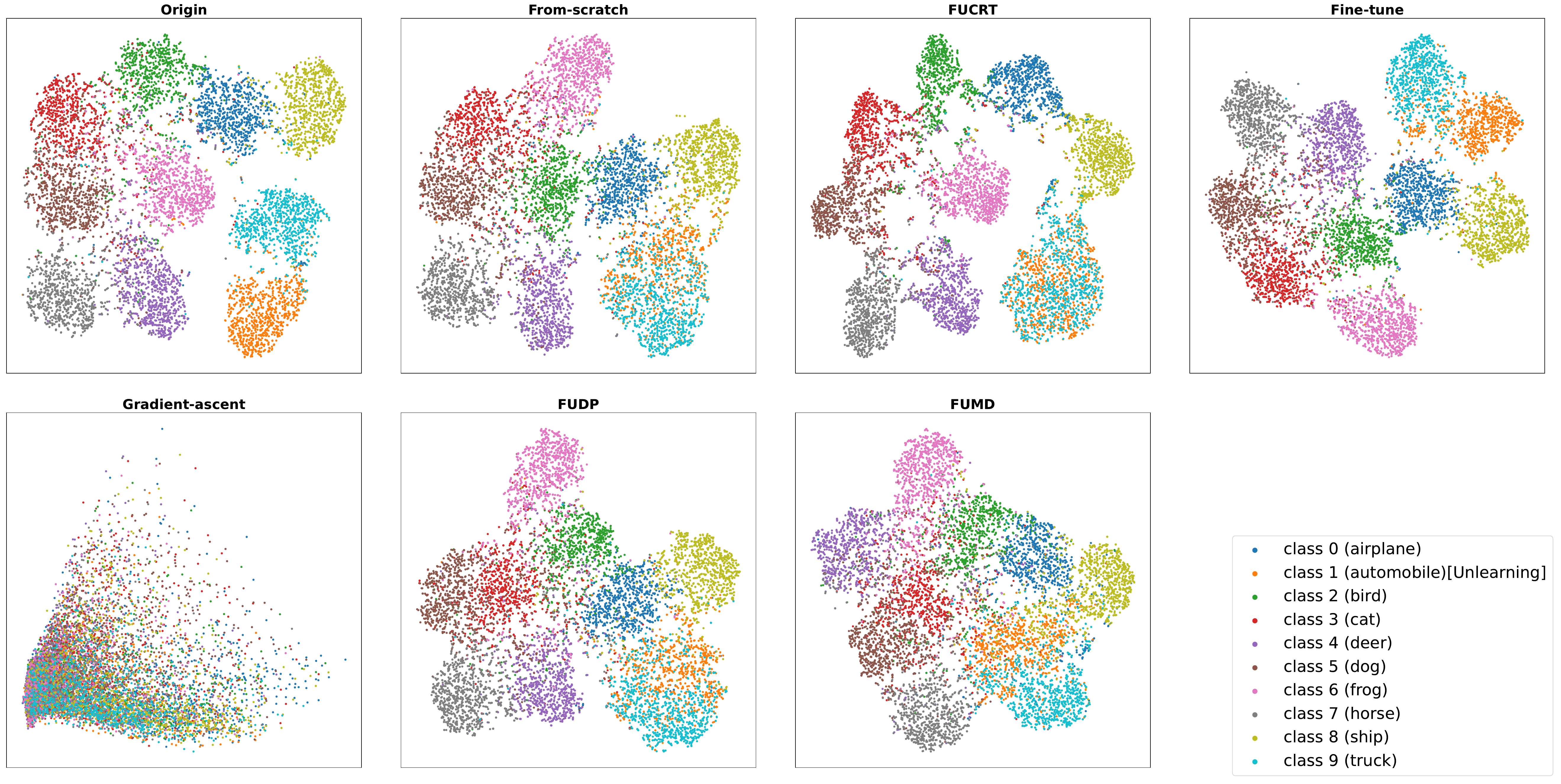}
  \caption{t-SNE visualization of the representation space distribution of unlearning methods with 10\% of the classes forgotten on CIFAR10 with the Non-IID setting.}
  \label{Fig:tsne2}
\end{figure*}

To better illustrate the differences in the representation space across various unlearning methods, we employed t-SNE to visualize the representation space of baselines and FUCRT after training on CIFAR10 under both IID and Non-IID settings. As shown in \figurename ~\ref{Fig:tsne1} and~\ref{Fig:tsne2}, points of different colors represent samples from different classes, with the orange points denoting the unlearning class (“automobile” class in CIFAR10).

Compared to the original model, the From-scratch method shows that the orange cluster has re-integrated into the adjacent clusters in the representation space, particularly the indigo cluster representing the “truck” class. Meanwhile, the clusters of the remaining classes are restructured within the representation space. This pattern was consistent across all four datasets, where the model predicts samples from unlearning classes as belonging to classes with similar features. The orange “automobile” class shares more features with the indigo “truck” class than with other classes (such as “bird”), leading to cluster merging in the representation space.

Our proposed FUCRT exhibits a similar characteristic as From-scratch method. Even after a few training rounds, the orange unlearning class is fully merged with the indigo transition class selected by the TCS strategy. Moreover, the remaining class clusters appear more compact compared to both the From-scratch and Origin models. This compactness is attributed to the transformation alignment, which optimizes the distribution of the representation space, amplifying inter-class differences while minimizing intra-class variance. This observation aligns with our experimental results, where FUCRT slightly outperforms the Origin and From-scratch methods on remaining class accuracy.

In the Fine-tune method's results, only a small portion of the unlearning class samples are distributed into neighboring clusters, and the unlearning class as a whole has not fully merged with the adjacent cluster. This is because fine-tuning focuses solely on the representation space of the remaining classes, leaving a significant amount of information from the unlearning classes untouched. On the other hand, the Gradient-ascent method severely disrupts the representation space, making it difficult to retrieve any meaningful information about either the remaining or unlearning classes.

FUDP and FUMD exhibit representation space behavior similar to the From-scratch method, with the unlearning class cluster merging with a nearby class cluster. However, comparing IID and Non-IID experimental results reveals that both methods are significantly affected by data distribution. Compared to From-scratch, the boundaries between remaining class clusters are less distinct, and the representation differences between different classes are less pronounced. In particular, FUMD's unlearning class additionally incorporates the originally distinguishable class (“ship”). Our proposed FUCRT specifically addresses federated unlearning in Non-IID settings. The transformation alignment effectively ensures consistency across the representation spaces of different clients. Experimental results demonstrate that our method maintains superior consistency in representation space distribution across IID and Non-IID settings.

\subsection{The Impact of Unlearning Proportion}
\label{sec:experimental.iup}

\begin{table}[t]
\caption{The impact of the unlearning proportion.}
\resizebox{0.48\textwidth}{!}{
\begin{tabular}{c|cccc|cccc}
\hline
\multirow{3}{*}{\textbf{\begin{tabular}[c]{@{}l@{}}Unlearning \\ Proportion\end{tabular}}} & \multicolumn{4}{c|}{\textbf{From-scratch}}                               & \multicolumn{4}{c}{\textbf{FUCRT}}                                      \\ \cline{2-9} 
                                                                                           & \multicolumn{2}{c}{\textbf{IID}} & \multicolumn{2}{c|}{\textbf{Non-IID}} & \multicolumn{2}{c}{\textbf{IID}} & \multicolumn{2}{c}{\textbf{Non-IID}} \\ \cline{2-9} 
                                                                                           & \textbf{ACC}    & \textbf{F1}    & \textbf{ACC}      & \textbf{F1}      & \textbf{ACC}    & \textbf{F1}    & \textbf{ACC}      & \textbf{F1}      \\ \hline
\textbf{10\%}                                                                              & 87.99           & 87.65          & 85.44             & 85.00            & 90.06           & 89.83          & 89.82             & 89.55            \\ \hline
\textbf{30\%}                                                                              & 93.09           & 92.87          & 91.27             & 90.93            & 92.77           & 92.68          & 92.10             & 91.98            \\ \hline
\textbf{50\%}                                                                              & 95.45           & 95.29          & 94.47             & 94.28            & 94.15           & 94.11          & 94.25             & 94.05            \\ \hline
\end{tabular}}
\label{table:unlearning_prop}
\end{table}

To evaluate FUCRT's efficacy with different unlearning proportions, we conducted experiments on the CIFAR10 dataset with unlearning proportions of 10\%, 30\%, and 50\%. As presented in \tablename ~\ref{table:unlearning_prop}, we compare FUCRT to the From-scratch method in terms of remaining class test accuracy (F1). We omit the accuracy (F1) results for the unlearning class, where both FUCRT and From-scratch achieve 0\%. This indicates that FUCRT provides 100\% erasure guarantee across different unlearning proportions, like the From-scratch method.

Our findings reveal that for both FUCRT and From-scratch methods, the accuracy and F1 of remaining classes increased as the proportion of unlearning classes grew. This trend can be attributed to the fact that fewer remaining classes result in a representation space that is easier to construct and optimize.

As previously shown, FUCRT outperforms retraining from scratch when 10\% of classes is forgotten. However, this advantage diminishes in more complex scenarios with higher unlearning proportions. This is primarily because FUCRT processes both remaining and unlearning class samples simultaneously during training, unlike From-scratch. With higher unlearning proportions, multiple classes may select the same transformation class, creating a larger transformation cluster. This can negatively impact other remaining classes by affecting the overall representation space structure.

Overall, FUCRT performs comparably to or better than From-scratch across different unlearning proportions.

\subsection{The Impact of Data Distribution}

\begin{figure}[tbp]
  \centering
  \includegraphics[width=0.495\textwidth]{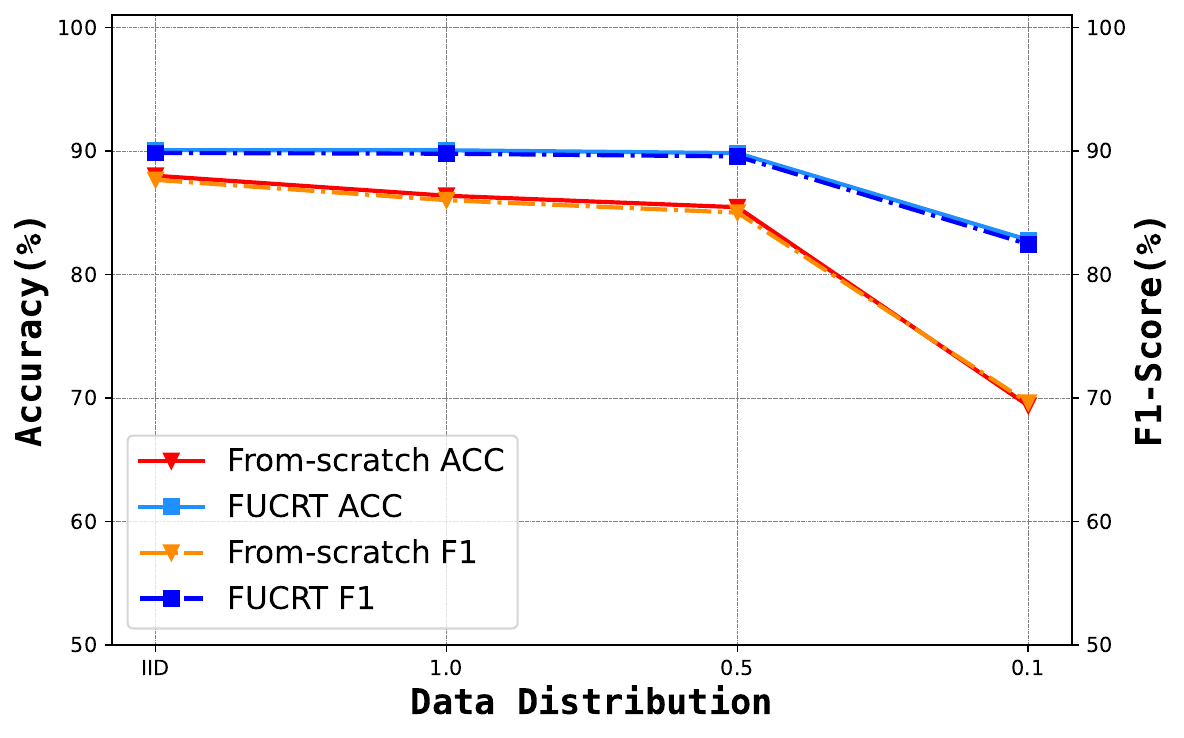}
  \caption{The impact of data distribution on remaining classes.}
  \label{Fig:data_distribution}
\end{figure}

To study the impact of data distribution on the performance of our method, we compared FUCRT and From-scratch under different data distributions, as illustrated in \figurename ~\ref{Fig:data_distribution}. We evaluated four scenarios with varying degrees of statistical heterogeneity: one IID setting and three Non-IID settings based on Dirichlet distributions. We plotted the remaining class accuracy and F1 of the unlearning models for each heterogeneity level. We omit the unlearning class results, where both FUCRT and From-scratch achieve 0\% accuracy (F1).

As shown in \figurename ~\ref{Fig:data_distribution}, FUCRT outperforms From-scratch in remaining class performance across all four data distribution scenarios. As statistical heterogeneity increases, differences in local representation space distributions among clients grow, which negatively impacts the performance of From-scratch on the remaining classes. Our method handles non-extreme heterogeneity scenarios well (IID and Dirichlet distributions with $\delta \geq 0.5$), where the unlearning model's performance on remaining classes doesn't decline significantly. This is because FUCRT's transformation alignment technique aligns local representation spaces across clients, reducing conflicts between class-aware representations. In extreme heterogeneity scenarios (Dirichlet distributions with $\delta=0.1$), our method alleviates the performance degradation of the remaining classes compared to the From-scratch method. It is noted that FUCRT's performance does decline due to the lack of majority class samples in high-heterogeneity settings. In other words, it is harder to guide the representation space distribution of other classes when clients train locally with few classes, weakening the effect of transformation alignment.

\subsection{Communication Cost}

\begin{table}[t]
\caption{ Communication Cost.}
\begin{tabular}{c|ccc}
\hline
\textbf{Dataset}  & \textbf{Prototype} & \textbf{Model} & \textbf{Ratio} \\ \hline
\textbf{CIFAR10}  & 5120               & 11,173,962     & 0.0004582      \\
\textbf{CIFAR100} & 51200              & 23,705,252     & 0.0021599      \\
\textbf{FMNIST}   & 5120               & 11,173,962     & 0.0004582      \\
\textbf{EuroSAT}  & 5120               & 11,173,962     & 0.0004582      \\ \hline
\end{tabular}
\label{table:communication}
\end{table}

To evaluate the communication cost of our FUCRT method, we present the amount of information each client needs to transmit across different datasets, as shown in \tablename ~\ref{table:communication}. In our method, each client computes a prototype for each class, represented as a 512-dimensional vector, and sends these prototypes to the server for aggregation. This means that, in addition to the model parameters exchanged during the standard federated learning process, clients also need to upload their local class prototypes.

However, the size of the class prototypes uploaded by each client is significantly smaller than that of the model parameters. For instance, when training a ResNet50 model on the CIFAR100 dataset, the model comprises 23,705,252 parameters, while the prototypes are calculated for 100 classes, each represented by a 512-dimensional vector, resulting in a total of 51,200 parameters for the prototypes. This accounts for only about 0.0021599 of the total model parameters. Similarly, for the CIFAR10, FMNIST, and EuroSAT datasets, the prototype sizes are even more negligible compared to the model sizes, constituting only about 0.0004582 of the total model parameters.

Overall, the additional communication overhead introduced by the class prototypes in FUCRT is minimal and acceptable.

\section{Conclusion}

In this paper, we have introduced Federated Unlearning via Class-aware Representation Transformation (FUCRT), a novel approach for achieving effective class-level unlearning in federated learning environments. FUCRT leverages class-aware representation transformation to identify optimal forgetting direction while employing dual class-aware contrastive learning to ensure consistent transformations across heterogeneous clients. Extensive experiments on multiple datasets demonstrate FUCRT's superior performance in terms of erasure guarantee, model utility preservation, and efficiency compared to state-of-the-art baselines. Our method achieves complete (100\%) erasure of unlearning classes while maintaining or even improving performance on remaining classes, and shows robustness to varying unlearning proportions and data distributions. Furthermore, FUCRT significantly reduces training rounds compared to retraining from scratch, with up to 40 times efficiency gains. By effectively addressing the challenges of unlearning in both IID and non-IID settings, FUCRT represents a significant step towards privacy-preserving federated learning systems. Future work may explore its applicability to feature-level unlearning scenarios.


\clearpage
\bibliographystyle{plain}
\bibliography{IEEETrans_bib}

\onecolumn
\appendix
\section{APPENDIX}

\subsection{Efficacy of 30\% Unlearning Proportion}

\begin{table*}[ht]

    \caption{ Accuracy and F1 on CIFAR10, CIFAR100, FMNIST, and EuroSAT datasets under the IID and Non-IID settings (\%). The unlearning class percentage is 30\%.}

    \centering

        \resizebox{0.9\textwidth}{!}{

    \begin{tabular}{c|c|cccc|cccc}

    \toprule

    \multirow{3}{*}{\textbf{Dataset}}   & \multirow{3}{*}{\textbf{Method}}       & \multicolumn{4}{c|}{\textbf{IID}}                                                 & \multicolumn{4}{c}{\textbf{Non-IID}}                                             \\ \cline{3-10}

                                        &                                        & \multicolumn{2}{c}{\textbf{Unlearning}} & \multicolumn{2}{c|}{\textbf{Remaining}} & \multicolumn{2}{c}{\textbf{Unlearning}} & \multicolumn{2}{c}{\textbf{Remaining}} \\ \cline{3-10}

                                        &                                        & ACC                          & F1       & ACC              & F1                   & ACC                & F1                 & ACC               & F1                 \\ \midrule

    \multirow{8}{*}{\textbf{CIFAR10}}  & Origin                                 & 84.33                        & 89.68    & 89.98            & 91.91                & 83.77              & 89.13              & 87.98             & 90.03              \\

                                        & From-scratch                           & 0.00                         & 0.00     & 93.09            & 92.87                & 0.00               & 0.00               & 91.27             & 90.93              \\

                                        & Fine-tune                              & 28.40                        & 45.16    & 94.73            & 94.60                & 7.29              & 21.94              & 94.98             & 94.75              \\

                                        & Gradient-ascent~\cite{wu2022federated} & 0.00                         & 0.00     & 18.34            & 27.64                & 0.00               & 0.00              & 28.82             & 37.66              \\

                                        & FUDP~\cite{wang2022federated}          & 0.00                         & 0.00     & 90.60            & 90.21                & 0.00               & 0.00               & 91.21             & 90.79              \\

                                        & FUMD~\cite{zhao2023federated}          & 0.00                         & 0.00     & 85.40            & 84.97                & 0.02               & 7.14               & 84.03             & 83.45              \\

                                        & \textbf{FUCRT} (Ours)                  & 0.00                         & 0.00     & 92.77            & 92.68                & 0.00               & 0.00               & 92.10             & 91.98              \\ \midrule

    \multirow{8}{*}{\textbf{CIFAR100}} & Origin                                 & 74.52                        & 87.88    & 76.27            & 84.44                & 75.22              & 88.87              & 75.73             & 84.12              \\

                                        & From-scratch                           & 0.00                         & 0.00     & 78.61            & 87.61                & 0.00               & 0.00               & 77.66             & 87.22              \\

                                        & Fine-tune                              & 50.13                        & 84.52    & 78.29            & 87.56                & 50.15              & 84.03              & 78.31             & 87.64              \\

                                        & Gradient-ascent~\cite{wu2022federated} & 0.00                         & 0.00     & 3.86             & 13.90                & 0.00               & 0.00               & 1.91              & 12.84              \\

                                        & FUDP~\cite{wang2022federated}          & 0.00                         & 5.00    & 74.11            & 80.45                & 5.87               & 68.59              & 71.46             & 78.06              \\

                                        & FUMD~\cite{zhao2023federated}          & 0.00                         & 0.00     & 6.27             & 42.94                & 0.00               & 0.00               & 8.58              & 44.88              \\

                                        & \textbf{FUCRT} (Ours)                  & 0.00                         & 0.00     & 75.43            & 85.88                & 0.00               & 0.00               & 74.99             & 85.62              \\ \midrule

    \multirow{8}{*}{\textbf{FMNIST}}    & Origin                                 & 91.96                        & 94.59    & 89.33            & 90.54                & 93.82              & 95.43              & 85.23             & 86.98              \\

                                        & From-scratch                           & 0.00                         & 0.00     & 93.74            & 93.58                & 0.00               & 0.00               & 91.02             & 90.40              \\

                                        & Fine-tune                              & 34.38                        & 71.49    & 93.58            & 93.40                & 31.19              & 60.89              & 92.66             & 92.36              \\

                                        & Gradient-ascent~\cite{wu2022federated} & 0.00                         & 0.00     & 10.60             & 19.10                & 0.00               & 0.00               & 16.24              & 25.61              \\

                                        & FUDP~\cite{wang2022federated}          & 0.00                         & 0.00     & 91.65            & 91.50                & 0.00               & 0.00               & 88.69             & 87.92              \\

                                        & FUMD~\cite{zhao2023federated}          & 0.00                         & 0.00     & 88.34            & 88.20                & 0.00               & 0.00               & 88.70             & 88.22              \\

                                        & \textbf{FUCRT} (Ours)                  & 0.00                         & 0.00     & 92.64            & 92.57                & 0.00               & 0.00               & 91.26             & 90.87              \\ \midrule

    \multirow{8}{*}{\textbf{EuroSAT}}   & Origin                                 & 94.98                        & 96.98    & 93.43             & 96.44                & 89.91              & 93.41              & 80.92             & 89.13              \\

                                        & From-scratch                           & 0.00                         & 0.00     & 95.95            & 97.94                & 0.00               & 0.00               & 87.14             & 92.76              \\

                                        & Fine-tune                              & 3.96                         & 9.85     & 97.41            & 98.69                & 0.93               & 5.35               & 94.71             & 97.28              \\

                                        & Gradient-ascent~\cite{wu2022federated} & 0.00                         & 0.00     & 18.86            & 47.46                & 0.00               & 0.00               & 31.48             & 73.64              \\

                                        & FUDP~\cite{wang2022federated}          & 0.00                         & 0.00     & 91.27            & 94.60                & 0.00               & 0.00               & 80.22             & 89.52              \\

                                        & FUMD~\cite{zhao2023federated}          & 0.00                         & 0.00     & 86.64            & 94.00                & 0.00               & 0.00               & 83.52             & 92.68              \\

                                        & \textbf{FUCRT} (Ours)                  & 0.00                         & 0.00     & 96.55            & 98.20                & 0.00               & 0.00               & 87.53             & 92.06              \\ \bottomrule

    \end{tabular}}

    \label{appendix.table:exp1}

    \end{table*}

To evaluate the performance of our proposed FUCRT more comprehensively under varying unlearning proportions, we present additional details on the performance of baselines and FUCRT under 30\% unlearning proportion across four datasets, as shown in \tablename ~\ref{appendix.table:exp1}. Similar to the setup in Section \ref{sec:experimental.egmup}, we use two metrics, Accuracy and F1-score, to evaluate the performance of the unlearning methods on both the unlearning and remaining classes.

Overall, as analyzed in Section \ref{sec:experimental.iup}, since fewer remaining classes need to be established and optimized in the new representation space, the remaining class accuracy of the unlearning methods slightly improves under 30\% unlearning proportion compared to 10\%. For instance, FUCRT’s remaining class accuracy increases by 1.59\% and 0.71\% under IID and Non-IID distributions, respectively. Fine-tune, however, continues to struggle in ensuring effective unlearning when a higher proportion of classes are unlearned; even under 30\% unlearning proportion, it still retains more than an accuracy of 3.96\% on the unlearning classes across the four datasets. On the contrary, Gradient-ascent fails to ensure the performance of the remaining classes, achieving a maximum remaining class accuracy of only 31.48\% across the four datasets.

FUDP maintains considerable effectiveness under 30\% unlearning proportion, with the remaining class accuracy in Non-IID settings across all four datasets showing an average decrease of 3.88\% compared to From-scratch (slightly higher than the 3.78\% observed under 10\% unlearning proportion). However, for another state-of-the-art method, FUMD, as the unlearning proportion increases, the gap in remaining class accuracy compared to From-scratch grows larger. For example, in the CIFAR10 Non-IID distribution, this gap widens from 2.84\% at 10\% unlearning proportion to 7.24\% at 30\%. This is primarily due to the increasing divergence between the degradation models trained by each client as the amount of unlearning data grows, which negatively impacts the final target model’s performance on the remaining classes.

Our proposed FUCRT continues to demonstrate outstanding performance under 30\% unlearning proportion. Across all datasets, FUCRT consistently maintains an accuracy of 0.00\% on the unlearning classes, ensuring complete erasure. Simultaneously, the remaining class performance of the model remains comparable to that of From-scratch, with only a 0.61\% decrease in remaining class accuracy across the four datasets. These results highlight that FUCRT outperforms other baselines in terms of overall performance under varying unlearning proportions.

\subsection{Representation Space Analysis for 30\% Unlearning Proportion}

\begin{figure*}[htbp]
  \centering
  \includegraphics[width=1.0\textwidth]{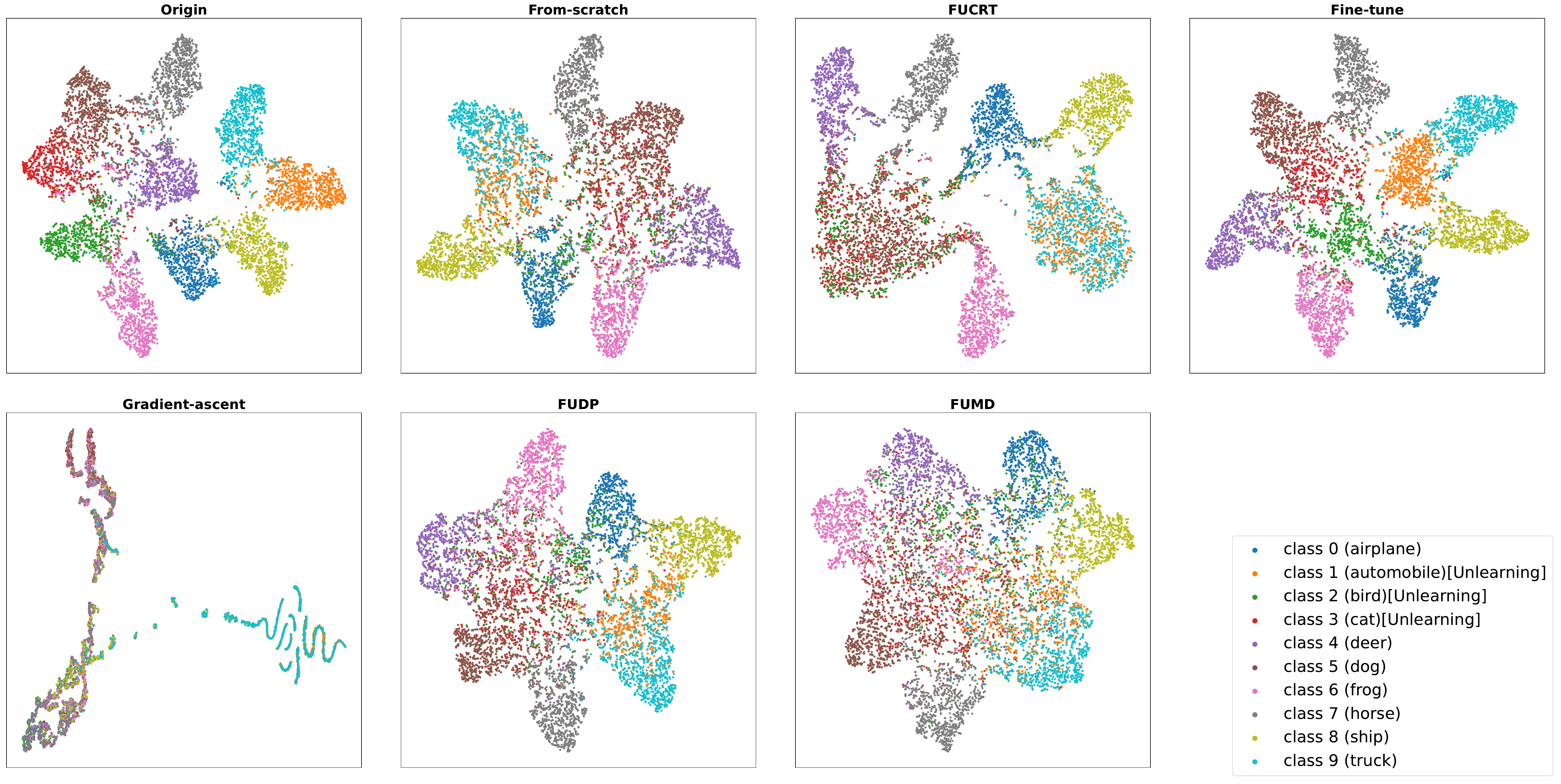}
  \caption{t-SNE visualization of the representation space distribution of unlearning methods with 30\% of the classes forgotten on CIFAR10 with the IID setting.}
  \label{appendix.Fig:tsne1}
\end{figure*}

\begin{figure*}[htbp]
  \centering
  \includegraphics[width=1.0\textwidth]{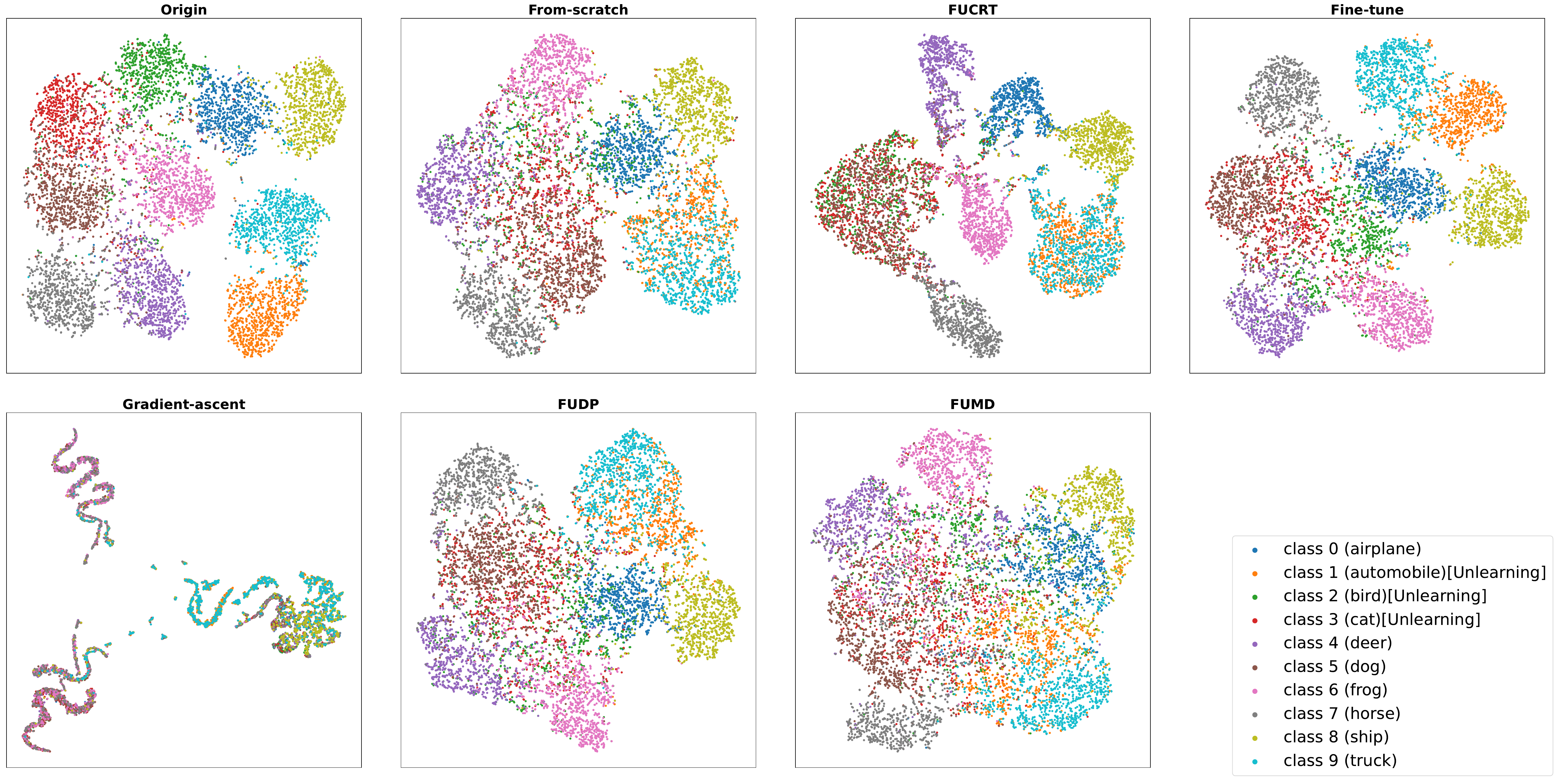}
  \caption{t-SNE visualization of the representation space distribution of unlearning methods with 30\% of the classes forgotten on CIFAR10 with the Non-IID setting.}
  \label{appendix.Fig:tsne2}
\end{figure*}

To further elucidate the distinctions among various unlearning methods under increased unlearning proportions, we present a comparative analysis of the representation spaces for baseline methods and FUCRT on the CIFAR10 dataset. This analysis encompasses both IID and Non-IID data distributions, with an unlearning proportion of 30\%, as illustrated in Figures \figurename ~\ref{appendix.Fig:tsne1} and \figurename ~\ref{appendix.Fig:tsne2}, respectively. In these visualizations, the unlearning classes for CIFAR10 are represented by orange points (“automobile” class), green points (“bird” class), and red points (“cat” class).

Examination of the t-SNE plot for the From-scratch method reveals notable changes compared to the original results. The orange “automobile” cluster has coalesced with the indigo “truck” cluster, a phenomenon consistent with observations from the 10\% unlearning proportion scenario. This merger can be attributed to the inherent similarity in features between these two classes. In contrast, the “bird” and “cat” clusters exhibit a more dispersed pattern, with data points scattered across multiple adjacent clusters and significant inter-cluster overlap. This dispersion is primarily due to the shared visual characteristics among various animal images, which naturally results in higher similarity across these classes.

As observed from the t-SNE plot of From-scratch, compared to the original results, the orange “automobile” cluster has merged with the indigo “truck” cluster, which is consistent with the results from 10\% unlearning proportion, as the features of these two classes are inherently more similar. In contrast, the “bird” cluster and “cat” cluster have multiple nearby clusters, causing their data points to scatter into different clusters and overlap with each other. This is primarily due to the fact that these different animal images share some common features, making them naturally more similar to one another.

In the representation space, our proposed FUCRT shows similar behavior with the orange cluster as it did under 10\% unlearning proportion. \figurename ~\ref{appendix.Fig:tsne1} and \figurename ~\ref{appendix.Fig:tsne2} illustrate that the “bird” and “cat” classes have been transferred to the “dog” class. These three classes form a cohesive and well-defined large cluster. In the context of the CIFAR10 dataset, our experiment involved forgetting 30\% of the classes, which equates to three classes. Theoretically, this should result in seven remaining classes, forming seven distinct clusters, as presented by the From-scratch method. Among all the baselines evaluated, only our proposed FUCRT consistently produced seven stable and clearly delineated clusters. This outcome aligns with the expected results based on the experimental design and the inherent structure of the CIFAR10 dataset.

Other baselines exhibit the same behavior under 30\% unlearning proportion as they did under 10\%. 
In the representation space of the Fine-tune method, while the unlearning classes are partially intermingled with the remaining classes, they remain distinctly separable and clustered according to their respective categories. This observation indicates that the Fine-tune approach's efficacy in erasing unlearning classes is suboptimal, as demonstrated in the experimental results presented in \tablename ~\ref{appendix.table:exp1}. Conversely, the Gradient-ascent method, although successful in achieving complete forgetting of the targeted classes through its gradient ascent technique, has significantly compromised the integrity of the representation space for the remaining classes. These findings underscore the inherent trade-off between effective unlearning and maintaining the quality of remaining information. Notably, as the unlearning proportion increases, FUDP and FUMD exhibit increased susceptibility to data heterogeneity. Compared to the IID setting, FUMD's decision boundaries become less distinct in Non-IID settings. In contrast, the transformation alignment in FUCRT ensures consistency in the distribution of representation spaces across different clients under Non-IID settings. The experimental results further demonstrate that our method maintains consistent representation space distributions across different unlearning proportions and data distributions.

\end{document}